\documentclass{article} 
\usepackage{iclr2027_conference,times}

\usepackage{amsmath,amsfonts,bm}

\def\eqref#1{equation~\ref{#1}}

\def\1{\bm{1}}

\DeclareMathAlphabet{\mathsfit}{\encodingdefault}{\sfdefault}{m}{sl}
\SetMathAlphabet{\mathsfit}{bold}{\encodingdefault}{\sfdefault}{bx}{n}

\newcommand{\E}{\mathbb{E}}

\usepackage{newtxtext}
\usepackage{amsmath,amssymb,amsthm}
\usepackage{mathtools}
\usepackage{bm}
\usepackage{graphicx}
\usepackage{booktabs}
\usepackage{multirow}
\usepackage{enumitem}
\usepackage{xcolor}
\usepackage{tikz}
\usetikzlibrary{arrows.meta,positioning,fit,backgrounds,calc}
\usepackage[expansion=false]{microtype}
\theoremstyle{definition}
\newtheorem{definition}{Definition}
\newtheorem{assumption}{Assumption}
\theoremstyle{plain}
\newtheorem{proposition}{Proposition}

\newcommand{\doop}{\mathrm{do}}
\newcommand{\Tcal}{\mathcal{T}}
\newcommand{\Pcal}{\mathcal{P}}
\newcommand{\Dcal}{\mathcal{D}}

\newcommand{\FrailPi}{\textsc{Frail-$\pi$}}
\newcommand{\FrailTau}{\textsc{Frail-$\tau$}}

\usepackage{hyperref}
\usepackage{url}

\usepackage{{natbib}}
\title{\centering Fragility Is Not Reducible to Hardness: \\Conditional Causal Difficulty for LLMs}

\author{\parbox{\textwidth}{\centering
Mohammad Reza Rezaei, Dingxian Wang, Nikhil Verma, \\
Sajjad Pakdamansavoji, Andrew Rabinovich \& Zhao Chen \\[8pt]
{\normalfont Upwork Inc. \\
Email: \texttt{mohammadrezarezaei@upwork.com}}}
}

\iclrfinalcopy 
\begin{document}

\maketitle
\lhead{Preprint}

\begin{abstract}
Large language models show large and seemingly arbitrary performance drops when a task is rewritten in ways that leave its requirements untouched. We recast task difficulty as a \emph{causal} quantity: the treatment effect of meaning-preserving perturbations on a model. A meaning-preserving rewrite can only be observed to hurt a task the model can already solve, so we place the estimand on the principal stratum where the degradation occurs: the \emph{conditional causal difficulty} (CCD), the probability that a randomly drawn twin breaks a solvable task. Because harder tasks tend to be more phrasing-sensitive, complexity enters as an effect modifier rather than a confounder, and we residualize it out to obtain \emph{complexity-residualized difficulty} (CRD); zero for a task exactly as fragile as its complexity predicts, positive for anomalously fragile ones. Current benchmarks do not supply the data CCD requires, so we introduce \textit{GlassGSM} and \textit{GlassMATH}, augmenting GSM8K and MATH with twins built from perturbation families, raising fragility on GlassGSM by $\approx 12\times$ over GSM-IC-style distractors on the same problems ($3.7\%\to44.9\%$) while keeping the answer key unchanged. Our predictor (\FrailPi) learns the per-perturbation break probability and reaches $0.613{\pm}0.023$ of the suite reliability ceiling on GlassGSM and $0.557{\pm}0.023$ on GlassMATH, against $0.171$/$0.401$ for a surface-feature baseline and $0.444$/$0.463$ for supervised baselines such as VAP3, and, at a $5\%$ curation budget on GlassGSM, doubles random recall of fragile items, matching an oracle ranking. Additionally, cross-fitted residualization leaves CRD uncorrelated with the declared proxy in practice. Code is available at \url{https://github.com/upwork/Conditional_Causal_Difficulty_for_LLMs}.
\end{abstract}

\section{Introduction}
\label{sec:intro}
\begin{figure}[h]
\centering
\includegraphics[width=1\textwidth]{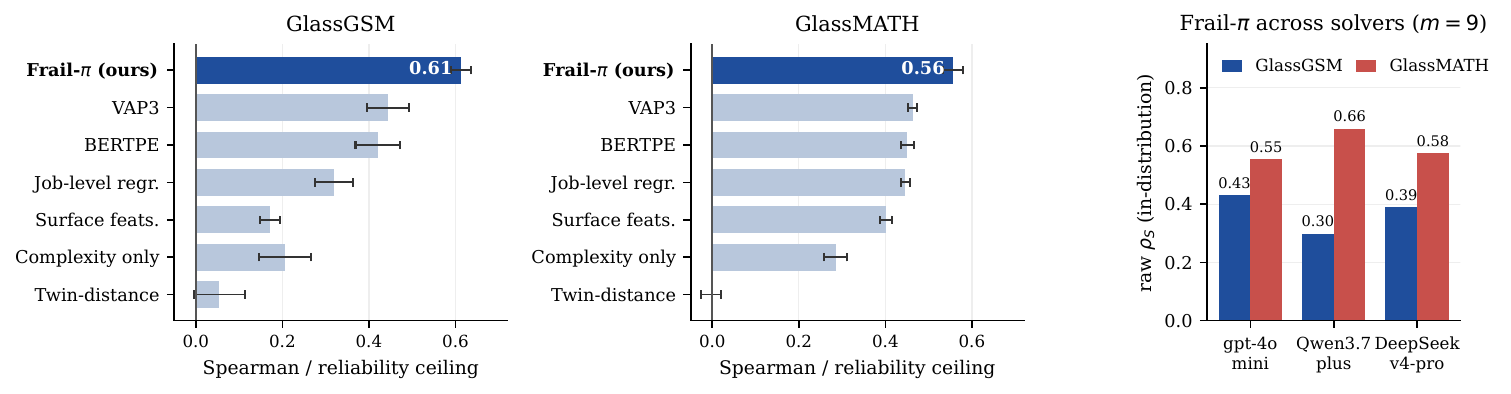}
\caption{\FrailPi\ predicts CCD better than every baseline on both suites (left, middle; Spearman normalized by the reliability ceiling, mean$\pm$sd over five seeds), and remains predictive when the solver backbone that defines the stratum is changed (right; raw $\rho_S$ of \FrailPi\ trained and evaluated on each solver's own CCD in the single-dose $m{=}9$ run, so not on the same scale as the left panels). Details in Tables~\ref{tab:c1} and~\ref{tab:solver-transfer}.}
\label{fig:teaser}
\end{figure}

\begin{figure}[t]
\centering
\begin{minipage}[t]{0.6\textwidth}
\centering
{\footnotesize\textbf{(a) Identical requirements, divergent scores}}\\[5pt]
\resizebox{\linewidth}{!}{%
\begin{tikzpicture}[
  task/.style={draw,rounded corners,fill=gray!8,align=left,font=\footnotesize,
               inner sep=5pt,text width=46mm},
  meas/.style={draw,rounded corners,fill=orange!10,align=center,font=\scriptsize,
               inner sep=4pt,text width=30mm},
  res/.style={draw,rounded corners,fill=green!10,align=center,font=\footnotesize,
              inner sep=5pt},
  arr/.style={-{Stealth[length=2mm]},semithick}
]
\node[task] (J) {%
  \textbf{Job $J$}\\[3pt]
  {\color{blue!55!black}\emph{A DVD can be played 1000}}\\
  {\color{blue!55!black}\emph{times before it breaks.}}\\
  \emph{There are two DVDs in the}\\
  \emph{public library, one has been}\\
  \emph{played 356 times and the}\\
  \emph{other has been played 135}\\
  \emph{times.}\\
  \emph{How many total times can}\\
  \emph{both DVDs be played}\\
  \emph{before they break?}%
};
\node[task,right=36mm of J] (TJ) {%
  \textbf{Twin $T(J)$}\\[3pt]
  \emph{There are two DVDs in the}\\
  \emph{public library, one has been}\\
  \emph{played 356 times and the}\\
  \emph{other has been played 135}\\
  \emph{times.}\\
  {\color{blue!55!black}\emph{A DVD can be played 1000}}\\
  {\color{blue!55!black}\emph{times before it breaks.}}\\
  \emph{How many total times can}\\
  \emph{both DVDs be played}\\
  \emph{before they break?}%
};
\draw[arr] (J.east) -- node[above,font=\scriptsize,align=center]
  {$T$: reorder two\\independent statements}
  (TJ.west);
\begin{scope}[on background layer]
\node[draw,dashed,rounded corners,fit=(J)(TJ),inner sep=3mm,
      label={[font=\scriptsize,inner sep=2pt]above:identical requirements}] (pair){};
\end{scope}
\node[meas,below=11mm of J] (scoreJ) {%
  model $+$ scorer $\phi$\\[3pt]$Y(J)=1.00$};
\node[meas,below=11mm of TJ] (scoreTJ) {%
  model $+$ scorer $\phi$\\[3pt]$Y(T(J))=0.00$};
\draw[arr] (J) -- (scoreJ);
\draw[arr] (TJ) -- (scoreTJ);
\coordinate (mid) at ($(scoreJ.south)!0.5!(scoreTJ.south)$);
\node[res,below=9mm of mid] (diff) {%
  \textbf{Observed contrast:}\quad
  $\Delta(J,T)\;=\;Y(J)-Y(T(J))\;=\;1.00-0.00\;=\;\mathbf{1.00}$%
};
\draw[arr] (scoreJ.south) -- (diff.north west);
\draw[arr] (scoreTJ.south) -- (diff.north east);
\end{tikzpicture}%
}
\end{minipage}%
\hfill%
\begin{minipage}[t]{0.40\textwidth}
\centering
{\footnotesize\textbf{(b) What the observational baselines estimate} (\S\ref{sec:background})}\\[5pt]
\resizebox{0.85\linewidth}{!}{%
\begin{tikzpicture}[
  var/.style={draw,circle,inner sep=1pt,minimum size=7mm,font=\small},
  cau/.style={-{Stealth[length=1.6mm]},semithick},
  conf/.style={-{Stealth[length=1.6mm]},semithick,dashed,red!70!black}
]
\node[var,label={[font=\scriptsize]above:complexity}] (C) {$C$};
\node[var,below left=9mm and 8mm of C,
      label={[font=\scriptsize]below:surface}] (S) {$S$};
\node[var,below right=9mm and 8mm of C,
      label={[font=\scriptsize]below:score}]   (Y) {$Y$};
\node[var,left=14mm of S,
      label={[font=\scriptsize]above:perturbation}] (T) {$T$};
\draw[conf] (C) -- (S);
\draw[conf] (C) -- (Y);
\draw[cau]  (S) -- (Y);
\draw[cau]  (T) -- (S);
\end{tikzpicture}%
}\\[4pt]
{\scriptsize
\textcolor{red!70!black}{\textbf{-\,-}}~back-door $S\!\leftarrow\!C\!\to\!Y$
(observational data only)
\quad\textbf{---}~causal path
\quad$T$ drawn independently of $C$: identified by design, not by adjustment}
\end{minipage}
\caption{\emph{(a)}~A real pair from \textbf{GlassGSM} ($m{=}1$): the twin $T(J)$ differs from $J$ only by moving the {\color{blue!55!black}blue} statement past an independent one. The model answers $J$ correctly but this twin incorrectly, so exact-match scoring gives $Y(J)=1.00$ and $Y(T(J))=0.00$, an instance of $\Delta(J,T)$ with no requirement-level basis. \emph{(b)}~The graph shows why $C$ confounds an \emph{observational} comparison across prompts through the back-door $S\leftarrow C\to Y$ (dashed), so any predictor of $Y(J)$ on a fixed phrasing estimates a complexity-inflated quantity. Our contrast does not open that back-door: $T$ is drawn independently of $C$, so comparing $J$ against its own twin is identified.}
\label{fig:overview}
\end{figure}
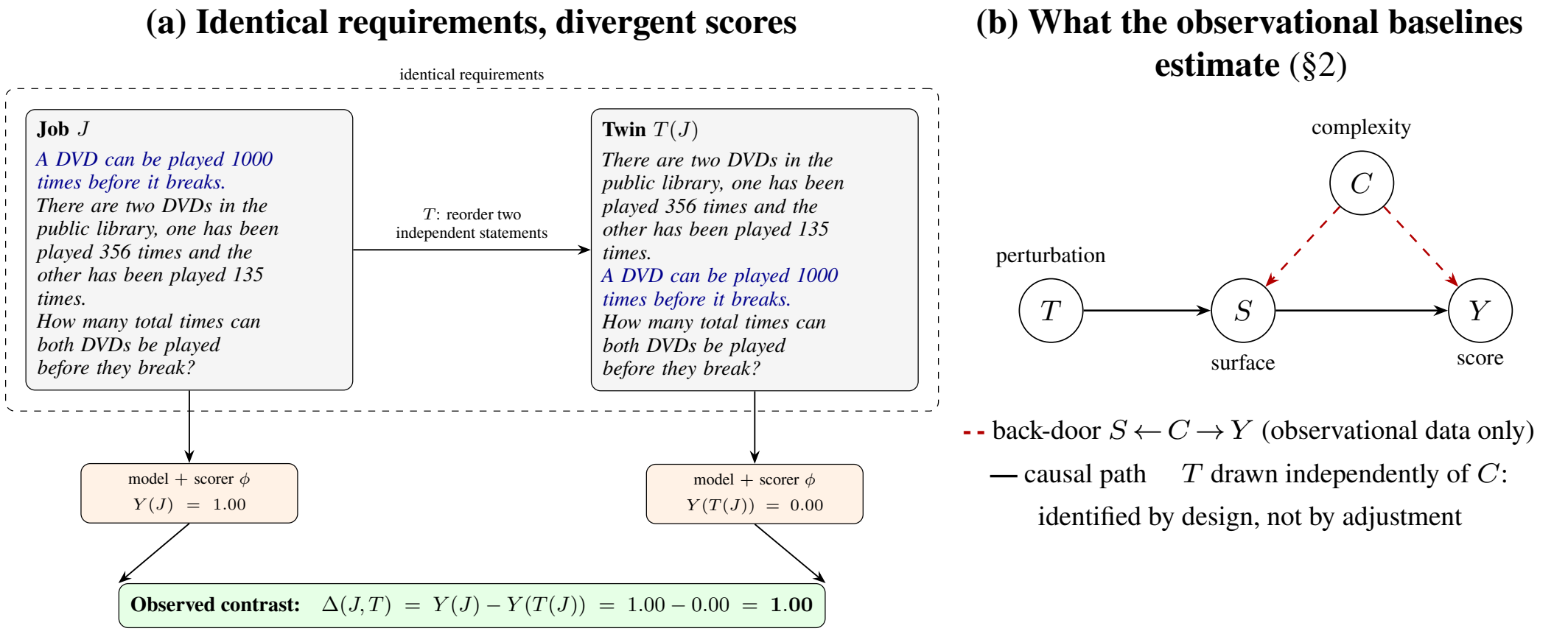

When a large language model performs well on a benchmark, that success implicitly promises the same performance on any semantically equivalent form. This promise often fails: two prompts with identical requirements and the same gold answer, differing only by the order of two logically independent statements, can receive very different scores (Figure~\ref{fig:overview}(a)); the reordering adds, removes, and renames nothing, yet the score moves anyway.

Most prior work either catalogs this brittleness descriptively
\citep{gsm-symbolic,promptbench,fragile-reasoning,prosa} or predicts the absolute performance of a task $J$, $Y(J)$, on a fixed phrasing \citep{irt-llm,amortized-eval,promptset,qpp-specificity,vap3,bertpe}; none of these estimates the \emph{causal magnitude} of a rewrite, forecasts it ahead of time, or factors out the part of sensitivity driven by intrinsic complexity. We therefore treat fragility as a \emph{causal} quantity: the effect of altering surface form while holding requirements constant. Identification here can be achieved by comparing $J$ with its own counterpart in a randomized experiment within-items without adjustment for complexity $C$. The back-door shown in Figure~\ref{fig:overview}(b), $S\leftarrow C\to Y$, characterizes what the \emph{alternatives} estimate: observational predictors $Y(J)$ and prior causal treatments of prompting, which compare between items and therefore must adjust for a proxy-recoverable confounder \citep{cpo,pryzant2021,veitch2020}. The causal framework instead does its work by fixing the estimand, settling \emph{where} it is defined, and providing the one adjustment the pairing does not make for free: $C$ is an \emph{effect modifier} (harder tasks tend to be more phrasing-sensitive), which is why we residualize it into CRD, zero for a task exactly as fragile as the typical task at its complexity level.

A meaning-preserving rewrite can only be seen to hurt a task the model already solves. On the complementary stratum ($Y(J){=}0$) the contrast is still observed, $\Delta(J,T)=-Y(T(J))\le0$, but it can only record a \emph{repair}, never a break. The unconditional average is therefore well defined but answers a different question: it nets breaks on solved tasks against repairs on unsolved ones, so its sign is set by base correctness rather than by how fragile a success is, and we find empirically that it is a far noisier target on GlassGSM (see Appendix~\ref{app:c1-davg-real}). Since our objective is to identify fragile successes, we present the conditional causal difficulty (CCD): the treatment effect on the solvable stratum, equivalently the fraction of the twin family that breaks a task. Measuring CCD requires perturbations that are meaning-preserving, fragile often enough to supply signal, and not inversely tied to complexity. Existing suites fail at least one requirement (\S\ref{sec:benchmarks}): GSM-Symbolic \citep{gsm-symbolic} moves the correct answer, and GSM-IC-style distractors \citep{gsm-ic} leave under $4\%$ of GSM8K test problems fragile. We therefore build \textbf{GlassGSM}, whose twins span three operator families at five dose levels. 

\textbf{Contributions.} \textbf{(1)} We formalize CCD as a principal-stratum estimand \citep{frangakis2002}, together with its logistic dose response (\S\ref{sec:new-estimand}), and show why the unconditional average $\tau{=}D_{\mathrm{avg}}$ is the wrong target (Appendix~\ref{app:c1-davg-real}). \textbf{(2)} We predict CCD by learning the per-perturbation break probability, which supplies up to $25$ supervised observations per task (across different twins at the five doses) in addition to a job-level scalar; its single-tower head predicts $\pi$ ex-ante, its per-twin predictions across doses give the dose-response parameters $(a_J,b_J)$, and residualizing $\pi$ gives the CRD, orthogonal to every function of a \emph{declared} complexity observable via cross-fitted residualization; this predictor outperforms all baselines on CCD on both suites and remains predictive across solver backbones (Figure~\ref{fig:teaser}). \textbf{(3)} We show the standard suites each fail one of three requirements and propose \textbf{GlassGSM} and \textbf{GlassMATH}, which raise fragility while preserving each task's answer key (the No-Op property, \S\ref{sec:benchmarks}).

\section{Background and Problem Setup}
\label{sec:background}
A prompt (\emph{task}; we use \emph{job} interchangeably) is a sequence of tokens $J\in\Tcal$ that an LLM maps to another sequence of output tokens, the response $O(J)$, which a \emph{quality-scoring function} $\phi$ collapses to a scalar
\begin{equation}
Y(J)\;=\;\phi\!\big(O(J),J\big)\;\in\;[0,1],
\label{eq:judge}
\end{equation}
an approximately unbiased measurement of real output quality. A
\emph{meaning-preserving perturbation operator} (MPPO) $T:\Tcal\to\Tcal$ rewrites a task into a variant with the same requirements but a different surface form (a cosmetic change); we call $T(J)$ the \emph{twin} of $J$ and let $\Pcal$ be a distribution over such operators. Three latent attributes recur below: intrinsic complexity $C$, requirements $R$, and surface form $S$; an MPPO acts on $S$ and leaves $R$ fixed by construction (Assumption~\ref{ass:invariance}).

\begin{assumption}[Requirement invariance]
\label{ass:invariance}
For every $T$ in the support of $\Pcal$ and every $J$: The scoring criteria of $\phi$ for $J$ and $T(J)$ are identical; hence an ideal solver attains $Y(J)=Y(T(J))$, and $T$ acts only on the surface form $S$; the requirements $R(J)=R(T(J))$ are left unchanged.
\end{assumption}

Under the potential-outcomes framework \citep{rubin1974,imbens2015causal},
each task has two scores, $Y(J)$ (control) and $Y(T(J))$ (treatment), with per-instance effect and conditional average treatment effect (CATE)
\begin{equation}
\Delta(J,T)\;=\;Y(J)-Y\!\big(T(J)\big),
\qquad
\tau(J)\;=\;\E_{T\sim\Pcal}\big[\,\Delta(J,T)\mid J\,\big],
\label{eq:ite-cate}
\end{equation}
valid under Assumption~\ref{ass:invariance} (the task-perturbation analogue of the prompt CATE of \citet{cpo}).

$T$ is drawn independently of the latent attributes of $J$, and $\Delta(J,T)$ compares the \emph{same} task under its original and a randomly assigned surface realization: a randomized experiment within the item, whose contrast is unbiased without any adjustment. The
back-door $S\leftarrow C\to Y$ \citep{pearl2009} enters instead as a description of what the \emph{alternatives} estimate: across two naturally occurring prompts $s\neq s'$, $C$ shapes both description and performance, so $P(Y\mid S{=}s)$ (the object every $Y(J)$ predictor in \S\ref{sec:related} regresses) is complexity-inflated. Given so permissive a design, the causal framework earns its place downstream, in three steps the pairing does \emph{not} give for free:
it fixes the \emph{estimand}, to be predicted ex-ante rather than measured
post-hoc (whence the measured base pass rate ranks our target \emph{negatively},
Table~\ref{tab:c1}); it forces the question of \emph{where} that estimand is
defined, a stratum problem \citep{frangakis2002} rather than a confounding one (\S\ref{sec:new-estimand}); and it demands one real adjustment, for effect modification by $C$ (\S\ref{sec:deconfounded}).

\section{Conditional causal difficulty (CCD)}
\label{sec:new-estimand}

IRT-style difficulty assigns a task a static score from average-model
performance \citep{irt-llm}, conflating hardness with sensitivity to phrasing. Causal difficulty separates them: zero whenever $J$ and its twin are equally hard, non-zero only when phrasing itself moves the score.

\begin{definition}[Unconditional causal difficulty (UCD)]
\label{def:difficulty}
The UCD of a task $J$ is
\begin{equation}
D_{\mathrm{avg}}(J)=\E_{T\sim\Pcal}\big[\Delta(J,T)\big]=\tau(J),
\end{equation}
the signed expected degradation of a typical innocuous edit, estimated from the $G$ twins sampled for $J$ ($G$ may vary by task).
\end{definition}

Definition~\ref{def:difficulty} applies to every task, including those the model fails outright, where $Y(J){=}0$ gives $\Delta(J,T)=-Y(T(J))\le0$ for every $T$ (and likewise for the worst case $D_{\mathrm{wc}}(J)=\max_T\Delta(J,T)$): there a twin can repair the task but never break it. The population average is thus well defined, but it is a net effect that mixes breaks on solved tasks with repairs on unsolved ones, not a measure of how fragile the model's successes are. Appendix~\ref{app:c1-davg-real} shows it is also a much noisier target --- scored against UCD, every method is weak (best $\rho_S{=}0.162$, a surface-length baseline). We therefore restrict the estimand to the \emph{solvable stratum} ($Y(J){=}1$), using the per-twin \emph{break event} $B(J,T)=\mathbf{1}\{Y(J){=}1,\ Y(T(J)){=}0\}$: a twin that breaks a task the model solved in that same measurement. On this stratum, with binary $Y$ (exact match on the real suites), $\Delta(J,T)=B(J,T)$ exactly, so the conditional fragility below is not a substitute quantity but the
treatment effect of~\eqref{eq:ite-cate} evaluated where a break can occur.

\begin{definition}[Conditional causal difficulty and dose response]
\label{def:new-estimands}
For a task $J$ with $Y(J){=}1$ and a family indexed by perturbation
\emph{dose} $m$ (here, one of five perturbation levels, $m\in\{1,3,5,7,9\}$),
\begin{equation}
\pi(J)=\E_{T\sim\Pcal}\big[B(J,T)\mid Y(J){=}1\big],
\qquad
\mathrm{logit}\,\Pr\big[B(J,T){=}1\mid m\big]=a_J+b_J\,(m-\bar m).
\end{equation}
$\pi(J)$ is the \emph{conditional causal difficulty} (CCD), also called the conditional fragility: the fraction of the family that breaks $J$. $a_J$ is fragility at typical dose and $b_J$ is the \emph{degradation rate}, how fast fragility grows with dose.
\end{definition}

Conditioning an effect on a potential outcome is the setting of \emph{principal stratification} \citep{frangakis2002}, which names CCD more precisely than a ``conditional CATE'' does. Taking the outcome as the stratifying variable, a (task, twin) pair falls in one of four basic strata indexed by $\big(Y(J),Y(T(J))\big)$ (always-solved, breaks, repaired, never-solved) and $\Delta{=}B$ on $\{Y(J){=}1\}$ says that $\pi(J)=\Pr[(1,0)\mid Y(J){=}1]$, the share of the \emph{solvable} union (always-solved plus breaks) that breaks. Our conditioning set is that union, defined by the control potential outcome alone, not a single stratum defined by the joint pair. Two features of the paired design make this the tractable case. Stratum membership is \emph{observed}: canonically only one arm is realized per unit, so strata are latent and principal effects need monotonicity or partial-identification bounds~\citep{zhang2003}, whereas here base and twins are all run, $Y(J)$ is measured for every item, and $\pi$ is estimated directly. And the conditioning opens no selection path: $T$ is drawn independently of every attribute of $J$, $Y(J)$ included, so restricting to $\{Y(J){=}1\}$ induces no collider bias through $T$; the remaining selection effect, regression to the mean from conditioning on a measured $Y(J)$, is measured by the identity-twin floor (Appendix~\ref{app:placebo}). What the restriction does cost (that $D_{\mathrm{avg}}\le\pi$ on average, that the partition is solver-relative, and that the stratum label is measured rather than given) is stated in \S\ref{sec:discussion}. For bookkeeping, we write $\mu_J(s)=\E[Y\mid J,\doop(S{=}s)]$ for task $J$'s response under a surface realization $s$, with $s=S(J)$ and $s'=S(T(J))$; the within-item contrast of \S\ref{sec:background} has conditional mean
\begin{equation}
\E\big[\Delta(J,T)\mid J,\,S{=}s,\,S'{=}s'\big]=\mu_J(s)-\mu_J(s'),
\label{eq:paired-id}
\end{equation}
with $J$ held fixed by the pairing, so no attribute of $J$ ($C$ included) need be observed or assumed unchanged by $T$.

\subsection{Complexity-residualized difficulty (CRD)}
\label{sec:deconfounded}
Decomposing
$\mu_J(s)=\alpha(C(J))+\beta(C(J),s)$ in \eqref{eq:paired-id}, the paired
contrast equals $\beta(C(J),s)-\beta(C(J),s')$, which still depends on $C(J)$
unless $\beta$ is additively separable: under effect modification the twin
returns a CATE at the task's own complexity, and the same holds verbatim for
$\pi$. Partialling $C$ out is therefore not an optional refinement but the
mechanism that separates anomalous fragility from intrinsic complexity, which
makes Definition~\ref{def:dperp} an important object.

\begin{definition}[Complexity-residualized difficulty (CRD)]
\label{def:dperp}
Fix a base quantity $Q\in\{\tau,\pi\}$, a \emph{declared} complexity observable
$C$, and a task population $p$ (restricted to the stratum
$Y(J){=}1$ when $Q=\pi$, since $\pi$ is only defined there). With
$\gamma_{Q,C,p}(c)=\E_{J'\sim p}[Q(J')\mid C(J')=c]$ the mean of $Q$ at $c$,
\begin{equation}
D^{\perp}_{Q,C,p}(J)\;=\;Q(J)-\gamma_{Q,C,p}\big(C(J)\big).
\label{eq:dperp}
\end{equation}
Our primary instantiation is $Q=\pi$, the estimand of
Definition~\ref{def:new-estimands}; residualizing it gives the CRD evaluated in
\S\ref{sec:c2}, which isolates anomalous fragility.
\end{definition}

CRD, $D^{\perp}$, is zero for a task exactly as fragile as the typical same-$C$ task and positive for \emph{anomalously} fragile tasks. Whereas $Q$ itself inherits the complexity of the task, CRD is the part the declared $C$ cannot account for.

\begin{proposition}[Residualization by projection]
\label{prop:orthogonality}
\textbf{(i)} CRD is the $L_2$ projection residual of $Q$ onto
measurable functions of the \emph{declared} $C$; hence $\E[D^{\perp}]=0$ and
$\E[D^{\perp}\,\eta(C)]=0$ for every $\eta\in L_2(\sigma(C))$, so CRD is
uncorrelated with \emph{every} function of $C$, for either $Q=\tau$ or $Q=\pi$.
Orthogonality to latent intrinsic complexity is \emph{not} implied, and holds
only insofar as $C$ is a sufficient summary of it.
\textbf{(ii)} CRD is estimated by cross-fitted residualization: regress
the measured $\hat Q$ on a complexity proxy out-of-fold to obtain $\hat\gamma$ and
set $\hat D^{\perp}=\hat Q-\hat\gamma$, consistent whenever $\hat Q\to Q$ and the
complexity regression is consistent \citep{dml,cpo} (Proof in Appendix~\ref{app:proof}).
\end{proposition}

\subsection{A complexity-residualized predictor}
\label{sec:predict}

Computing the estimand exactly requires running the model on $J$ and every twin from
$\Pcal$, which might be prohibitive at scale. We instead learn a predictor that takes the text of $J$ and is \emph{causally aligned}: it must estimate the effect
of a surface change, not learn that verbose descriptions score low, so we make
the \emph{contrast} the architectural primitive. A single shared encoder
$f_\theta$ embeds $J$ and $T(J)$ into $\bm{h}=f_\theta(J)$ and
$\bm{h}'=f_\theta(T(J))$; their difference $\bm{h}-\bm{h}'$ feeds a contrast head
$\psi_\rho$, while a single-tower head $g_\omega$ predicts difficulty when no
twin exists at acquisition time:
\begin{equation}
\widehat\Delta(J,T)=\psi_\rho\big(\bm{h}-\bm{h}'\big),
\qquad
\hat D(J)=g_\omega\big(\bm{h}\big).
\label{eq:heads}
\end{equation}
The residualized prediction subtracts a cross-fitted complexity regression from the single-tower output, $\widehat D^{\perp}(J)=\widehat D(J)-\hat\gamma(C(J))$
(Prop.~\ref{prop:orthogonality}(ii)); this shares the additive term $-\hat\gamma(C(J))$ with the target, and \S\ref{sec:c2} confirms the resulting prediction is near-orthogonal to $C$ in practice, not just the measured target (Table~\ref{tab:c2}), holding across suites and across generator models.
Our deployed instantiation is the $Q{=}\pi$ predictor \FrailPi\ (\S\ref{sec:c1}), which applies this recipe to the per-observation break event $B(J,T)$ and so receives up to $25$ labelled contrasts per task ($m$ twins at each of the five doses $m\in\{1,3,5,7,9\}$, pooled from the released dose files). The same recipe applied to the unconditional $\tau$ (\FrailTau) is retained for Synth-CD, where scores are continuous and per-twin effects are non-negative by construction, so $\tau$ contains no repairs and measures degradation directly (\S\ref{sec:c2}, Appendices~\ref{app:synth2} and~\ref{app:c1-davg-real}).

\paragraph{Training objective.}
We \emph{amortize} the inner expectation of \eqref{eq:ite-cate} so difficulty for an unseen task costs one forward pass, minimizing $\E_{J\sim p_{\Dcal}}[\mathcal{L}(\theta,\rho,\omega;J)]$ with per-task term
\begin{equation}
\mathcal{L}(\cdot;J_i)=\sum_{T\in\mathcal{S}_i}
   \ell_\delta\!\big(\widehat\Delta(J_i,T),\Delta(J_i,T)\big)
 +\ell_\delta\!\big(g_\omega(\bm{h}_i),\,
   \widehat D^{\,\mathrm{obs}}(J_i)\big)
 +\lambda_{\mathrm{inv}}\,\mathcal{R}_{\mathrm{inv}}(\theta;J_i),
\label{eq:loss}
\end{equation}
Here, $\mathcal{S}_i$ is the set of collected twins of $J_i$, $\widehat D^{\,\mathrm{obs}}(J_i)$ is its measured job-level target, and $\ell_\delta$ is matched to the target. For \FrailPi\ on the real suites the targets are binary break events (on the solvable stratum $\Delta(J_i,T)=B(J_i,T)\in\{0,1\}$) and break rates in $[0,1]$, so $\ell_\delta$ is the binary cross-entropy on a sigmoid output; cross-entropy is a proper scoring rule, so the contrast term is minimized in expectation by $\Pr[B(J,T){=}1\mid J,T]$ and the job-level term by the task's break rate, an estimate of $\pi(J)$; the reported $\hat\pi$ is the single-tower output $g_\omega(\bm{h})$, which needs no twin. For \FrailTau\ (Synth-CD and Appendix~\ref{app:c1-davg-real}), whose targets are signed effects rather than probabilities, $\ell_\delta$ is the Huber loss with threshold $\delta$. The first (contrast) term supplies per-observation causal supervision to the shared encoder, and the second fits the deployable single-tower head to the observed rate. The encoder decomposes
$\bm{h}=c(C(J))+v(R(J),S(J))$ into a complexity part $c(\cdot)$ and a surface part $v(\cdot)$; assuming, for this regularizer only (identification in \eqref{eq:paired-id} does not need it), that $T$ leaves $C$ as well as $R$ fixed,
\begin{equation}
\bm{h}-\bm{h}'\;=\;v\!\big(R(J),S(J)\big)-v\!\big(R(J),S(T(J))\big).
\label{eq:cancel}
\end{equation}
Since $c(C(J))$ cancels, giving a contrast that depends on $S$ alone, we can encourage the decomposition with an invariance regularizer, $\mathcal{R}_{\mathrm{inv}}(\theta)=\E_{J,T\sim\Pcal}\|\Pi_C\bm{h}-\Pi_C\bm{h}'\|^2$, with $\Pi_C$ a learned complexity-subspace projection.

\section{Benchmarking causal difficulty}
\label{sec:benchmarks}

To estimate CCD, a benchmark must meet three conditions: (\textbf{R1}, No-Op) the perturbation satisfies Assumption~\ref{ass:invariance}, so a score drop reflects fragility and not a harder problem; (\textbf{R2}, signal) enough problems are actually fragile to support a ranking metric; and (\textbf{R3}, no inverse confound) fragility is not driven \emph{down} by intrinsic complexity, which would leave the target nearly unlearnable. Table~\ref{tab:fragility-comparison} scores widely used public suites against these criteria.

\begin{table}[t]
\centering\small
\caption{Where each real-math perturbation suite fails. \textit{Fragile}: $d_{\mathrm{obs}}>0.05$, with $d_{\mathrm{obs}}$ the measured UCD $D_{\mathrm{avg}}$ over all problems (this table only; elsewhere, e.g.\ Appendix~\ref{app:utility}, a job is fragile when $\pi>0.05$ on the solvable stratum); \textit{Zero}: $|d_{\mathrm{obs}}|<0.001$. \textbf{R1}~No-Op (meaning preserving); \textbf{R2}~enough fragile problems to rank; \textbf{R3}~no inverse complexity--fragility confound. \textit{MATH (original)} has no perturbation suite, so Fragile/Zero, R1, and R3 are n/a. GSM-IC~\citep{gsm-ic} releases 100 base problems; the GSM-IC row applies its irrelevant-sentence construction to all 1{,}319 GSM8K test problems (20 twins each), the base set GlassGSM also uses. For GSM-Plus, answer preservation was verified empirically (zero drift across the full test set). GlassGSM and GlassMATH numbers are for the released corpora pooled over all five doses (unique twins), resampled from the audited pools of Appendix~\ref{app:noop-audit}.}
\label{tab:fragility-comparison}
\begin{tabular}{@{}lccccccc@{}}
\toprule
Dataset & Problems & Twins & Fragile & Zero & R1 & R2 & R3 \\
\midrule
GSM-IC (on GSM8K test) & 1{,}319 & 20 & 3.7\%  & 90.1\% & \checkmark & $\times$ & \checkmark \\
GSM-Plus (MPPO)~\citep{gsm-plus}
                   & 1{,}319 & 2  & 10.5\% & 86.2\%
                   & \checkmark & $\times$ & \checkmark \\
GSM-Symbolic p1    & 100     & 10 & 25.0\% & 56.0\% & $\times$ & \checkmark & n/a \\
GSM-DC \citep{gsm-dc-shi-2025} & 6{,}222 & 1  & 16.4\% & 64.0\% & \checkmark & \checkmark & $\times$ \\
\midrule
\textbf{GlassGSM (ours)}
                   & \textbf{1{,}319} & \textbf{avg.\ 13.9} & \textbf{44.9\%} & \textbf{48.6\%}
                   & \checkmark & \checkmark & \checkmark \\
MATH (original)~\citep{hendrycks-math}
                   & 5{,}000 & --- & n/a & n/a
                   & n/a & $\times$ & n/a \\
GlassMATH (ours)
                   & 4{,}999 & avg.\ 23.1 & 36.5\% & 34.4\%
                   & \checkmark & \checkmark & \checkmark \\
\bottomrule
\end{tabular}
\end{table}

Existing suites fail at least one requirement. \emph{GSM-Symbolic} \citep{gsm-symbolic} moves the correct answer, violating \textbf{R1}. \emph{GSM-DC} \citep{gsm-dc-shi-2025} passes R1/R2 but fails \textbf{R3}: its measured fragility falls as complexity rises. \emph{GSM-IC}~\citep{gsm-ic} (its construction applied to GSM8K test) and the original \emph{MATH}~\citep{hendrycks-math} both fail \textbf{R2}: GSM-IC's inert filler is ignored ($3.7\%$ fragile), and MATH ships with no perturbation suite at all. \emph{GSM-Plus}~\citep{gsm-plus} passes \textbf{R1}/\textbf{R3} on its two MPPOs (problem understanding, distraction insertion) but fails \textbf{R2}: fragility is $10.5\%$, too thin for reliable per-instance ranking. We include it in Table~\ref{tab:fragility-comparison} to complete the survey.

We therefore introduce \textbf{GlassGSM}, whose twins are drawn from three operator families: entity renaming, sentence reordering, and injecting a \emph{math-domain} distractor sentence from a \emph{different} GSM8K problem; we generate twins across all three families at five doses $m\in\{1,3,5,7,9\}$, with higher doses injecting more distractor sentences. The contamination audit (Appendix~\ref{app:noop-audit}) removes twins whose injected distractor sentence accidentally reproduces the target problem's answer key (the dominant leakage mode we screened for); the released GlassGSM corpus, resampled from the audited pools and pooled over all five doses, holds $18{,}309$ unique twins ($32{,}858$ pairs) over its $1{,}319$ problems (avg.\ $13.9$ twins/problem). \textbf{GlassMATH}
applies the same three families, doses, and audit criterion to Hendrycks MATH~\citep{hendrycks-math}, additionally carrying each item's MATH difficulty level (1--5, shipped with the dataset) as a complexity proxy that is external to our pipeline (set independently of any operator draw or model output, hence non-circular), and its released corpus, pooled over the five doses, holds $115{,}364$ unique twins ($122{,}847$ pairs) over $4{,}999$ problems (avg.\ $23.1$ twins/problem). We release both suites, with all five dose files, in public repositories (linked in the abstract); GlassGSM is hosted on the Hugging Face Hub at \url{https://huggingface.co/datasets/reza2ei/GlassGSM}.

\section{Experiments}
\label{sec:experiments}

We evaluate our method on Synth-CD (known causal ground truth) and the real math data GlassGSM/GlassMATH. \textbf{C1} (\S\ref{sec:c1}): the CCD is predictable from task text ahead of time, while the UCD on the same suite is not (Appendix~\ref{app:c1-davg-real}). \textbf{C2} (\S\ref{sec:c2}): the CRD is orthogonal to the declared complexity observable, and Appendix~\ref{app:utility} shows the ranking
is actionable under a curation budget. Baselines: surface features, a declared-complexity score, a job-level regression on frozen TF-IDF+SVD features, frozen twin-distance, three PromptSET measures \citep{promptset,qpp-specificity}, measured base pass rate $\bar Y(J)$,
BERTPE~\citep{bertpe}, and VAP3~\citep{vap3}. The declared-complexity score is a regressor on several complexity features; CRD is residualized on $C$ alone, so this score can still rank CRD (Figure~\ref{fig:cross-model}). Unless stated otherwise, twins are generated and solved by \texttt{gpt-4o-mini}; the solver decodes at temperature $0.2$ with $5$ samples per prompt, and $Y$ scores the modal answer, so a twin could break a task by re-sampling alone. An identity-twin control (Appendix~\ref{app:placebo}) measures this floor: re-solving unperturbed stratum bases breaks $1.1\%$ (GlassGSM) and $2.5\%$ (GlassMATH) of re-solves, and the modal base answer changes for at most $0.3\%$ of items, so stratum membership is stable. Every pipeline run that builds a group of twins also re-solves the base, so each task has several base measurements: $\bar Y(J)$ is the fraction of correct base samples ($5$ per run) over all runs, a task enters the solvable stratum when $\bar Y(J)\ge0.5$ (the operational form of $Y(J){=}1$ in every result below), and $B(J,T)$ counts a break only in runs where the base was solved. $\bar Y(J)$ is therefore not constant on the stratum (it is below $1$ whenever any base sample is wrong, even when the modal answer is stable), and this is what the base-correctness row of Table~\ref{tab:c1} ranks. Backbone specs and the training recipe are in Appendix~\ref{app:impl}; the datasets in Appendix~\ref{app:datasets}; UCD results on Synth-CD and GlassGSM in Appendices~\ref{app:synth2} and~\ref{app:c1-davg-real}.

\subsection{C1: CCD is predictable from text}
\label{sec:c1}

Our main model, \FrailPi, fine-tunes a BERT-base encoder end to end with the break-probability head, evaluated against the baselines with 5 different seeds. Table~\ref{tab:c1} reports CCD and the dose-response parameters, with every Spearman normalized by that suite's reliability ceiling ($0.969$ and $0.972$; Appendix~\ref{app:audit}) so $1.0$ is the attainable maximum. \FrailPi\ reaches $0.613{\pm}0.023$ of ceiling on GlassGSM ($\rho_S{=}0.594$) and $0.557{\pm}0.023$ on GlassMATH ($\rho_S{=}0.541$), well above the surface-feature and declared-complexity-score baselines. Representation choice (TF-IDF+SVD, frozen sentence/OpenAI encoders) is ablated in Appendix~\ref{app:c1-ablations}: fully fine-tuning the encoder beats every frozen representation on both suites.

\emph{TF-IDF+SVD features, job-level regression} trails \FrailPi\ on both suites, but it differs from \FrailPi\ in both the encoder (frozen) and the objective (job-level), so on its own it does not say which change matters. Holding the per-observation objective fixed, fine-tuning beats every frozen representation ($0.613$ vs.\ ${\le}0.398$ on GlassGSM, $0.557$ vs.\ ${\le}0.435$ on GlassMATH; Table~\ref{tab:c1-repr}), so fine-tuning accounts for much of the lift. With TF-IDF+SVD features held fixed, the per-observation objective adds $0.078$ on GlassGSM ($0.398$ vs.\ $0.320$) and nothing on GlassMATH, where it is slightly lower ($0.435$ vs.\ $0.446$; Tables~\ref{tab:c1} and~\ref{tab:c1-repr}). Five further training-recipe levers (Appendix~\ref{app:c1-ablations}) do not close the remaining gap to the ceiling either.

Scored against the UCD on the GlassGSM holdout (unsolved tasks included), every method, ours included, is weak (best $\rho_S{=}0.162$, surface length, vs.\ \FrailPi's $0.594$ on CCD; Appendix~\ref{app:c1-davg-real}): restricting to the solvable stratum (\S\ref{sec:new-estimand}) is what turns a net break-minus-repair average into a rankable target.

\begin{table}[tb]
\centering\small
\caption{Test-fold Spearman \emph{normalized by the suite reliability ceiling} ($0.969$ for GlassGSM, $0.972$ for GlassMATH; Appendix~\ref{app:audit}), mean${\pm}$sd over five seeds. $\pi$: CCD predicted by the single-tower head $g_\omega$; $a$/$b$: dose-response intercept/slope (Definition~\ref{def:new-estimands}), from a logistic fit to the contrast head's per-twin predictions across doses (uses the twins). \emph{Job-level regression}: frozen TF-IDF+SVD features regressed directly on job-level $\pi$; it differs from \FrailPi\ in both encoder and objective (\S\ref{sec:c1}). Frozen twin-distance / Prompt Clarity / Embedding Entropy / Embedding Specificity-IEF (from PromptSET, \citealt{promptset,qpp-specificity}) are retargeted to $\pi$ with a TF-IDF+SVD embedding on the same seeded splits (Appendix~\ref{app:extended-baselines}); they have no per-dose grid, so $a$/$b$ are n/a. The $a$/$b$ columns use the same $\pi$ ceiling as a common scale, not ceilings of their own.}
\label{tab:c1}
\resizebox{\textwidth}{!}{%
\begin{tabular}{@{}llccc@{}}
\toprule
Suite & Method & $\pi$ $\uparrow$ & $a$ $\uparrow$ & $b$ $\uparrow$ \\
\midrule
\multirow{11}{*}{\shortstack[l]{GlassGSM}}
 & \textbf{\FrailPi\ }         & \textbf{0.613${\pm}$0.023} & \textbf{0.611${\pm}$0.020} & \textbf{0.289${\pm}$0.021} \\
 & VAP3                                             & 0.444${\pm}$0.049 & 0.441${\pm}$0.046 & 0.245${\pm}$0.031 \\
 & BERTPE                                           & 0.420${\pm}$0.051 & 0.420${\pm}$0.049 & 0.255${\pm}$0.035 \\
 & TF-IDF+SVD features, job-level regr.            & 0.320${\pm}$0.044 & 0.320${\pm}$0.049 & 0.184${\pm}$0.019 \\
 & Surface features (GBT)                           & 0.171${\pm}$0.023 & 0.164${\pm}$0.026 & 0.146${\pm}$0.056 \\
 & Declared complexity score only                   & 0.207${\pm}$0.060 & 0.192${\pm}$0.058 & 0.181${\pm}$0.049 \\
 & Frozen twin-distance                             & 0.054${\pm}$0.059 & n/a & n/a \\
 & Prompt Clarity (PromptSET)                       & $-$0.060${\pm}$0.035 & n/a & n/a \\
 & Embedding Entropy (PromptSET)                    & 0.023${\pm}$0.026 & n/a & n/a \\
 & Embedding Specificity-IEF (PromptSET)            & 0.034${\pm}$0.033 & n/a & n/a \\
 & Measured base pass rate $\bar Y(J)$              & $-$0.124${\pm}$0.037 & $-$0.130${\pm}$0.032 & 0.011${\pm}$0.017 \\
\midrule
\multirow{11}{*}{\shortstack[l]{GlassMATH}}
 & \textbf{\FrailPi\ }         & \textbf{0.557${\pm}$0.023} & \textbf{0.548${\pm}$0.022} & \textbf{0.492${\pm}$0.015} \\
 & VAP3                                             & 0.463${\pm}$0.010 & 0.454${\pm}$0.010 & 0.172${\pm}$0.026 \\
 & BERTPE                                           & 0.451${\pm}$0.015 & 0.441${\pm}$0.017 & 0.272${\pm}$0.053 \\
 & TF-IDF+SVD features, job-level regr.            & 0.446${\pm}$0.010 & 0.431${\pm}$0.008 & 0.182${\pm}$0.021 \\
 & Surface features (GBT)                           & 0.401${\pm}$0.014 & 0.389${\pm}$0.015 & 0.180${\pm}$0.025 \\
 & Declared complexity score only                   & 0.285${\pm}$0.026 & 0.299${\pm}$0.028 & 0.242${\pm}$0.044 \\
 & Frozen twin-distance                             & $-$0.003${\pm}$0.024 & n/a & n/a \\
 & Prompt Clarity (PromptSET)                       & 0.003${\pm}$0.027 & n/a & n/a \\
 & Embedding Entropy (PromptSET)                    & $-$0.160${\pm}$0.032 & n/a & n/a \\
 & Embedding Specificity-IEF (PromptSET)            & 0.020${\pm}$0.025 & n/a & n/a \\
 & Measured base pass rate $\bar Y(J)$              & $-$0.285${\pm}$0.040 & $-$0.280${\pm}$0.042 & 0.057${\pm}$0.036 \\
\bottomrule
\end{tabular}%
}

\end{table}

\subsection{C2: CRD is recoverable and orthogonal}
\label{sec:c2}

On the real suites the primary C2 result uses $Q=\pi$
(Definition~\ref{def:new-estimands}), residualized via \FrailPi\ against $C$. Table~\ref{tab:c2} (single-dose $m{=}9$ run) shows orthogonality holds on both suites for the measured
target itself ($|$Corr(target,$C$)$|{\le}0.06$, as
Prop.~\ref{prop:orthogonality}(i) predicts for a consistent complexity regression) and for the model's prediction (Corr(pred,$C$) similarly near zero), as Prop.~\ref{prop:orthogonality}(ii) predicts once the cross-fitted complexity regression is consistent.

Synth-CD has continuous scores and non-negative per-twin effects, so $\tau$ contains no repairs and measures degradation directly there, and the suite isolates a question the real suites cannot: whether the contrast architecture is
\emph{intrinsically} complexity-orthogonal, independent of any explicit residualization. Both analytic factors are calibrated to be strong effect modifiers ($C\perp E$ by construction), so there is a large complexity-explained
component to remove; the \FrailTau\ contrast head reaches $\rho_S{=}0.617$ at $|\rho_C|{=}0.028$ (Appendix~\ref{app:synth2}), roughly twentyfold below the raw
$D_{\mathrm{avg}}$-level $|\rho_C|{\approx}0.60$, and gets there \emph{without} explicit residualization, since differencing cancels the complexity component of the representation (Eq.~\ref{eq:cancel}), leaving the head to rank mainly on the $C$-independent
expertise signal $E$; cf.\ Table~\ref{tab:c2}.
\begin{figure}[ht]
\centering
\includegraphics[width=\textwidth]{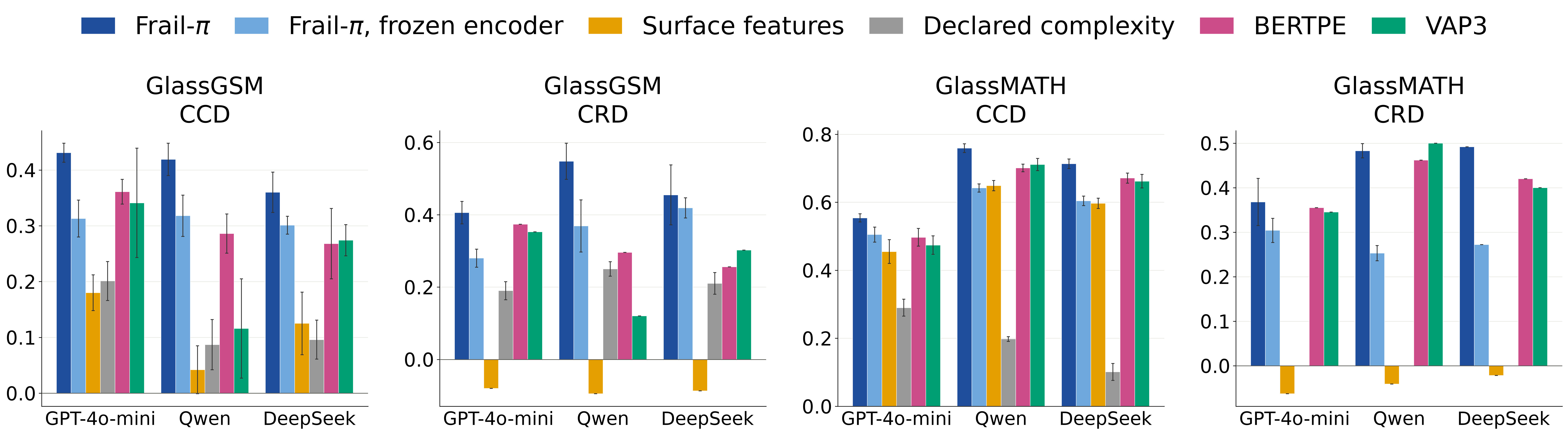}
\caption{Cross-model generalization, single-dose ($m{=}9$) twins, mean$\pm$sd over five seeds (sd${<}0.005$ for bars without a visible error bar). Qwen and DeepSeek denote the Qwen3.7-plus and DeepSeek-v4-pro generators. The \emph{\FrailPi} bars are the cross-generator run whose orthogonality diagnostics appear in Table~\ref{tab:c2}; shown alongside two ablations not in that table: a frozen-encoder representation (\emph{\FrailPi, frozen encoder}) and per-generator retrained BERTPE/VAP3. Other baselines follow Table~\ref{tab:c1}'s definitions. Bars are raw (unnormalized) $\rho_S$. This single-dose slice differs from the pooled 5-dose run behind Table~\ref{tab:c1}, so its numbers are not directly comparable to that table's.}
\label{fig:cross-model}
\end{figure}

\textbf{Generalization across generator models.} To check whether \FrailPi's ranking ability is specific to that generator, we reran the fine-tuned \FrailPi\ on single-dose ($m{=}9$) twins generated from the same base problems by Qwen3.7-plus and DeepSeek-v4-pro (labelled Qwen and DeepSeek below), scoring \texttt{gpt-4o-mini} twins on the matching single-dose slice for a fair comparison (CCD ranking in Figure~\ref{fig:cross-model}; orthogonality diagnostics in Table~\ref{tab:c2}). CCD replicates for both generators (higher on GlassMATH; on GlassGSM comparable for Qwen and somewhat lower for DeepSeek), so the ranking is not specific to one generator; the residualized CRD stays orthogonal to $C$ for both generators on both suites (Table~\ref{tab:c2}). On the GlassMATH CRD panel BERTPE and VAP3 come close to \FrailPi, and VAP3 is slightly ahead with Qwen twins, so the cross-generator advantage is clearest on CCD.
\begin{table}[ht]
\centering \small
\caption{CRD orthogonality against $C$; the primary instantiation of Definition~\ref{def:dperp}. The declared proxy $C$ is suite-specific: the MATH difficulty level (1--5) for GlassMATH (external, non-circular; \S\ref{sec:benchmarks}), and the number of reasoning steps in each problem's GSM8K gold solution for GlassGSM, which ships no such label. $|\rho_C|$: absolute Spearman between the prediction and $C$; $\mathrm{Corr}$: Pearson. All rows are \FrailPi\ on the single-dose ($m{=}9$) run of Figure~\ref{fig:cross-model}, differing only in the twin generator.}
\label{tab:c2}
\begin{tabular}{@{}lcccc c@{}}
\toprule
& & & \multicolumn{2}{c}{$\mathrm{Corr}(\cdot,C)$} & $\mathrm{Corr}(\cdot,C^2)$ \\
\cmidrule(lr){4-5}\cmidrule(lr){6-6}
Suite / generator & MAE$\downarrow$ & $|\rho_C|\downarrow$ & target & pred. & pred. \\
\midrule
GlassGSM, gpt-4o-mini twins & 0.129 & 0.044 & $-$0.043 & $-$0.061 & $-$0.068 \\
GlassMATH, gpt-4o-mini twins & 0.129 & 0.030 & $0.046$ & $-$0.023 & $-$0.022 \\
GlassGSM, Qwen3.7-plus twins & 0.074 & 0.092 & $-$0.058 & $-$0.055 & $-$0.069 \\
GlassMATH, Qwen3.7-plus twins & 0.117 & 0.067 & $-$0.018 & $-$0.016 & $-$0.012 \\
GlassGSM, DeepSeek-v4-pro twins & 0.058 & 0.087 & $-$0.025 & $-$0.060 & $-$0.056 \\
GlassMATH, DeepSeek-v4-pro twins & 0.103 & 0.053 & $-$0.055 & $-$0.040 & $-$0.018 \\
\bottomrule
\end{tabular}
\end{table}
\textbf{Solver transfer.} The checks above fix the solver (\texttt{gpt-4o-mini}); since the stratum is defined by the solver's own $Y(J)$, varying the \emph{solver} directly tests how far the estimand travels: on the same GlassGSM and GlassMATH items and twins ($m{=}9$), we independently measure \FrailPi's target --- each solver's own CCD --- for \texttt{gpt-4o-mini}, Qwen3.7-plus, and DeepSeek-v4-pro, train the fine-tuned \FrailPi\ on one solver's measured $\pi$, and evaluate the held-out ranking against a \emph{different} solver's measured $\pi$ on the same holdout problems (Table~\ref{tab:solver-transfer}). Off-diagonal $\rho_S$ is positive throughout, and several cells are comparable to one of the two solvers' own in-distribution accuracy (e.g.\ on GlassGSM, DeepSeek-v4-pro-trained transfer \emph{into} Qwen3.7-plus reaches $0.306$, above Qwen3.7-plus's own $0.299$; the \texttt{gpt-4o-mini}-trained model likewise beats the in-distribution models for Qwen3.7-plus, $0.355$ vs.\ $0.299$, and for DeepSeek-v4-pro, $0.411$ vs.\ $0.389$): a predictor trained on one solver's fragility labels ranks a different solver's fragility well above chance without ever seeing that solver's labels. Transfer is not
uniformly close to in-distribution accuracy, so we read this as partial, solver-dependent transfer rather than solver-invariance, as the solver-relativity of the stratification (\S\ref{sec:new-estimand}) predicts. We therefore evaluate transfer on CCD: CRD is residualized within one solver's stratum, so it is a within-solver tool.
\begin{table}[ht]
\centering\small
\caption{\textbf{Solver transfer.} Held-out $\rho_S$ of the fine-tuned
\FrailPi\ trained on the row solver's measured $\pi$ and
evaluated against the column solver's measured $\pi$, same items/twins
throughout ($m{=}9$). Diagonal = in-distribution; off-diagonal =
cross-solver transfer.}
\label{tab:solver-transfer}
\begin{tabular}{@{}lccc@{}}
\toprule
\multicolumn{4}{@{}l}{\emph{GlassGSM}} \\
train $\backslash$ eval & gpt-4o-mini & Qwen3.7-plus & DeepSeek-v4-pro \\
\midrule
gpt-4o-mini      & 0.431 & 0.355 & 0.411 \\
Qwen3.7-plus     & 0.297 & 0.299 & 0.401 \\
DeepSeek-v4-pro  & 0.237 & 0.306 & 0.389 \\
\addlinespace
\multicolumn{4}{@{}l}{\emph{GlassMATH}} \\
train $\backslash$ eval & gpt-4o-mini & Qwen3.7-plus & DeepSeek-v4-pro \\
\midrule
gpt-4o-mini      & 0.554 & 0.180 & 0.295 \\
Qwen3.7-plus     & 0.205 & 0.660 & 0.468 \\
DeepSeek-v4-pro  & 0.270 & 0.456 & 0.576 \\
\bottomrule
\end{tabular}
\end{table}
\section{Related Work}
\label{sec:related}
\textbf{Non-causal views.}
\citet{gsm-symbolic,promptbench} catalogue task fragility descriptively; \citet{fragile-reasoning,prosa} localize failures post-hoc or via decoding confidence. IRT-style difficulty \citep{irt-llm,atlas,easy2hard,amortized-eval,smart} assigns a static score from a single phrasing --- distinct from CCD: a task can be hard yet robust or easy yet fragile. Ours is a causal contrast on the solvable stratum, refined into the proxy-orthogonal CRD.
\textbf{Per-instance prompt performance prediction.}
\citet{promptset} predict per-prompt correctness via variation-interaction and embedding-specificity signals \citep{qpp-specificity}; \citet{vap3} add VAP3 and BERTPE~\citep{bertpe}. We retarget five of these --- Prompt Clarity, Embedding Entropy, Embedding Specificity-IEF, VAP3, and BERTPE --- to CCD (Definition~\ref{def:new-estimands}, Table~\ref{tab:c1}; adaptation in Appendix~\ref{app:extended-baselines}). \emph{How we differ:} all five were designed to predict $Y(J)$ on a fixed phrasing, not the contrast on the solvable stratum, so predicting $Y(J)$ is not the same as predicting which solvable problems break; indeed the measured base pass rate $\bar Y(J)$ ranks $\pi$ \emph{negatively} on both suites (Table~\ref{tab:c1}). None of them separates fragility from complexity $C$; our CRD does. VAP3, Prompt Clarity, and Embedding Entropy also use held-out variation text at test time (an oracle condition); our ex-ante head uses the base text and still beats all three (Table~\ref{tab:c1}).

\textbf{Causal views.}
\citet{robustness-causal} frame perturbation robustness causally; CPO estimates prompt CATEs by double machine learning \citep{cpo,dml}; other text-causal work adjusts for text-recoverable confounding
\citep{pryzant2021,veitch2020,counterfactual-baselines,text-interventions,gpi,catt};
\citet{contrastive-structured-treatments} identify causal latents via a contrastive structured-treatment representation. All compare \emph{across} items
and must recover a confounder from text; we perturb the \emph{task}, so identification is a property of the design, not of an adjustment (Eq.~\ref{eq:paired-id}), and we claim no contribution for it. \emph{How we
differ} lies downstream: $C$ enters as an effect \emph{modifier}, so CRD (the object of study, not a selection criterion) still demands a real adjustment. Closest in kind is \textbf{principal stratification} \citep{frangakis2002}: our estimand is a principal-stratum quantity (\S\ref{sec:new-estimand}) rather than a population average that nets breaks against repairs. That literature's defining difficulty (latent stratum membership, handled by monotonicity or partial identification \citep{zhang2003}) does not arise here, since the paired design realizes both arms on every item; what carries over is its interpretive discipline, that an effect defined on a stratum is a statement about that stratum and not about the population as a whole
(\S\ref{sec:discussion}).
\section{Conclusion}
\label{sec:discussion}

We recast task difficulty for LLMs as a causal measure on the principal stratum where a meaning-preserving rewrite can be observed to bite: the CCD, the effect of such a rewrite on tasks a fixed model already solves, predicted per instance and ahead of time. \FrailPi\ learns the per-perturbation break probability; its single-tower head predicts CCD ex-ante, its per-twin predictions across doses give the dose response, and residualizing CCD gives CRD; it beats every baseline and frozen representation we test on CCD and roughly doubles random recall of fragile GlassGSM items under a small curation budget, matching an oracle ranking at $5\%$, while the same sweep scored against the UCD is weak for every method. The causal framing fixes the estimand, its stratum, and the residualization separating intrinsic complexity from model-specific weakness, supporting difficulty-aware dataset design and targeted robustness assessment.

\textbf{Limitations.} The principal stratum (\S\ref{sec:new-estimand}) is solver-relative, and cross-solver transfer, though positive, falls below in-distribution accuracy in several cells (Table~\ref{tab:solver-transfer}), so the estimand is not exactly solver-invariant. The stratum label is a measured $Y(J)$, so $\pi$ carries a re-sampling floor, measured with identity twins ($1.1\%$ and $2.5\%$ of re-solves; Appendix~\ref{app:placebo}), and since $D_{\mathrm{avg}}\le\pi$ on average, $\pi$ describes solvable tasks, not the whole population. CRD is orthogonal only to the \emph{declared} proxy; a different $C$ would yield a different CRD (Prop.~\ref{prop:orthogonality}). The No-Op audit (Appendix~\ref{app:noop-audit}) is a best-effort filter, not a proof of zero contamination, though a second LLM adjudicator over all ${\approx}10$K flagged-but-kept GlassMATH pairs and human review of a $1{,}000$-pair sample confirmed every pair it kept as clean.

\section*{Use of Large Language Models}
\label{sec:llm-usage}

We used generative AI for three purposes. \textbf{Dataset construction (required).} The twins defining GlassGSM and GlassMATH are LLM-generated: an LLM added each
meaning-preserving perturbation (\S\ref{sec:benchmarks}, Appendix~\ref{app:benchmarks}), and the
cross-generator ablations vary which model does so. \textbf{LLM-as-judge for data validation.} The No-Op audit filtering those twins uses an LLM judge (\texttt{gpt-5}) to decide whether a twin changes or invalidates the gold answer, with a second, independently queried adjudicator (Claude) for the cross-checks in Appendix~\ref{app:noop-audit}; verdicts are combined with deterministic structural collision checks rather than trusted alone. \textbf{Writing.} We used LLM tools to edit and polish wording and formatting of text we wrote. LLMs were not used to generate ideas or hypotheses, develop the theory, formulate or prove the mathematical claims, design the methodology or experiments, or interpret results; the remaining
disclosure categories do not apply.

We have reviewed all AI-assisted work: every generated pair passes through the automated filter, every unit-collision-flagged GlassMATH pair was additionally re-adjudicated by a second LLM judge over the full ${\approx}10$K-pair pool, a $1{,}000$-pair stratified sample was verified by human review against the base text, twin text, and gold answer (confirming every kept pair clean; sampled LLM drops were conservative false positives), and all LLM-edited manuscript text was read and approved by the authors. We take
responsibility for the final content of this work, including text, claims, and artifacts produced with the aid of generative AI.

\bibliography{iclr2027_conference}

@INPROCEEDINGS{catt,
  author={Yang, Xu and Zhang, Hanwang and Qi, Guojun and Cai, Jianfei},
  booktitle={2021 IEEE/CVF Conference on Computer Vision and Pattern Recognition (CVPR)}, 
  title={Causal Attention for Vision-Language Tasks}, 
  year={2021},
  volume={},
  number={},
  pages={9842-9852},
  doi={10.1109/CVPR46437.2021.00972}}

@inproceedings{gsm-symbolic,
  title={Gsm-symbolic: Understanding the limitations of mathematical reasoning in large language models},
  author={Mirzadeh, Iman and Alizadeh-Vahid, Keivan and Shahrokhi, Hooman and Tuzel, Oncel and Bengio, Samy and Farajtabar, Mehrdad},
  booktitle={International Conference on Learning Representations},
  volume={2025},
  pages={94743--94765},
  year={2025}
}

@misc{cpo,
      title={Optimizing Prompts for Large Language Models: A Causal Approach}, 
      author={Wei Chen and Yanbin Fang and Shuran Fu and Fasheng Xu and Xuan Wei},
      year={2026},
      eprint={2602.01711},
      archivePrefix={arXiv},
      primaryClass={cs.AI},
      url={https://arxiv.org/abs/2602.01711}, 
}

@article{irt-llm,
  title={Learning compact representations of LLM abilities via item response theory},
  author={Chen, Jianhao and Wang, Chenxu and Zhang, Gengrui and Ye, Peng and Bai, Lei and Hu, Wei and Qu, Yuzhong and Hu, Shuyue},
  journal={arXiv preprint arXiv:2510.00844},
  year={2025}
}

@article{atlas,
  title={Adaptive testing for llm evaluation: A psychometric alternative to static benchmarks},
  author={Li, Peiyu and Tang, Xiuxiu and Chen, Si and Cheng, Ying and Metoyer, Ronald and Hua, Ting and Chawla, Nitesh V},
  journal={arXiv preprint arXiv:2511.04689},
  year={2025}
}

@article{counterfactual-baselines,
  title={Compared to What? Baselines and Metrics for Counterfactual Prompting},
  author={Yang, Zihao and Levy, Mosh and Goldberg, Yoav and Wallace, Byron C},
  journal={arXiv preprint arXiv:2605.01048},
  year={2026}
}

@inproceedings{robustness-causal,
    title = "Interpreting the Robustness of Neural {NLP} Models to Textual Perturbations",
    author = "Zhang, Yunxiang  and
      Pan, Liangming  and
      Tan, Samson  and
      Kan, Min-Yen",
    editor = "Muresan, Smaranda  and
      Nakov, Preslav  and
      Villavicencio, Aline",
    booktitle = "Findings of the Association for Computational Linguistics: ACL 2022",
    month = may,
    year = "2022",
    address = "Dublin, Ireland",
    publisher = "Association for Computational Linguistics",
    url = "https://aclanthology.org/2022.findings-acl.315/",
    doi = "10.18653/v1/2022.findings-acl.315",
    pages = "3993--4007"
}

@inproceedings{text-interventions,
    title = "Estimating Causal Effects of Text Interventions Leveraging {LLM}s",
    author = "Guo, Siyi  and
      Marmarelis, Myrl G  and
      Morstatter, Fred  and
      Lerman, Kristina",
    editor = "Inui, Kentaro  and
      Sakti, Sakriani  and
      Wang, Haofen  and
      Wong, Derek F.  and
      Bhattacharyya, Pushpak  and
      Banerjee, Biplab  and
      Ekbal, Asif  and
      Chakraborty, Tanmoy  and
      Singh, Dhirendra Pratap",
    booktitle = "Proceedings of the 14th International Joint Conference on Natural Language Processing and the 4th Conference of the Asia-Pacific Chapter of the Association for Computational Linguistics",
    month = dec,
    year = "2025",
    address = "Mumbai, India",
    publisher = "The Asian Federation of Natural Language Processing and The Association for Computational Linguistics",
    url = "https://aclanthology.org/2025.findings-ijcnlp.10/",
    doi = "10.18653/v1/2025.findings-ijcnlp.10",
    pages = "170--190",
    ISBN = "979-8-89176-303-6"
}

@article{dml,
    author = {Chernozhukov, Victor and Chetverikov, Denis and Demirer, Mert and Duflo, Esther and Hansen, Christian and Newey, Whitney and Robins, James},
    title = {Double/debiased machine learning for treatment and structural parameters},
    journal = {The Econometrics Journal},
    volume = {21},
    number = {1},
    pages = {C1-C68},
    year = {2018},
    month = {02},
    issn = {1368-4221},
    doi = {10.1111/ectj.12097},
    url = {https://doi.org/10.1111/ectj.12097},
    eprint = {https://academic.oup.com/ectj/article-pdf/21/1/C1/27684918/ectj00c1.pdf},
}

@book{pearl2009,
  author = {Pearl, Judea},
  edition = {2nd},
  publisher = {Cambridge University Press},
  title = {Causality: Models, Reasoning and Inference},
  year = 2009
}

@inproceedings{easy2hard,
author = {Ding, Mucong and Deng, Chenghao and Choo, Jocelyn and Wu, Zichu and Agrawal, Aakriti and Schwarzschild, Avi and Zhou, Tianyi and Goldstein, Tom and Langford, John and Anandkumar, Anima and Huang, Furong},
title = {Easy2Hard-Bench: standardized difficulty labels for profiling LLM performance and generalization},
year = {2024},
isbn = {9798331314385},
publisher = {Curran Associates Inc.},
address = {Red Hook, NY, USA},
booktitle = {Proceedings of the 38th International Conference on Neural Information Processing Systems},
articleno = {1407},
numpages = {43},
location = {Vancouver, BC, Canada},
series = {NIPS '24}
}

@article{fragile-reasoning,
  title={Fragile Reasoning: A Mechanistic Analysis of LLM Sensitivity to Meaning-Preserving Perturbations},
  author={Han, Shou-Tzu and Rizk, Rodrigue and Santosh, KC},
  journal={arXiv preprint arXiv:2604.01639},
  year={2026}
}

@misc{contrastive-structured-treatments,
      title={Contrastive representations of high-dimensional, structured treatments}, 
      author={Oriol Corcoll Andreu and Athanasios Vlontzos and Michael O'Riordan and Ciaran M. Gilligan-Lee},
      year={2024},
      eprint={2411.19245},
      archivePrefix={arXiv},
      primaryClass={stat.ML},
      url={https://arxiv.org/abs/2411.19245}, 
}

@article{gpi,
author = {Kosuke Imai and Kentaro Nakamura},
title = {Causal Inference with Generative Artificial Intelligence: Application to Texts as Treatments},
journal = {Journal of the American Statistical Association},
volume = {0},
number = {0},
pages = {1--14},
year = {2026},
publisher = {Taylor \& Francis},
doi = {10.1080/01621459.2026.2689629},


URL = { 
    
        https://doi.org/10.1080/01621459.2026.2689629
    
    

},
eprint = { 
    
        https://doi.org/10.1080/01621459.2026.2689629
    
    

}

}

@article{amortized-eval,
  title={Reliable and efficient amortized model-based evaluation},
  author={Truong, Sang and Tu, Yuheng and Liang, Percy and Li, Bo and Koyejo, Sanmi},
  journal={arXiv preprint arXiv:2503.13335},
  year={2025}
}

@inproceedings{prosa,
    title = "{P}ro{SA}: Assessing and Understanding the Prompt Sensitivity of {LLM}s",
    author = "Zhuo, Jingming  and
      Zhang, Songyang  and
      Fang, Xinyu  and
      Duan, Haodong  and
      Lin, Dahua  and
      Chen, Kai",
    editor = "Al-Onaizan, Yaser  and
      Bansal, Mohit  and
      Chen, Yun-Nung",
    booktitle = "Findings of the Association for Computational Linguistics: EMNLP 2024",
    month = nov,
    year = "2024",
    address = "Miami, Florida, USA",
    publisher = "Association for Computational Linguistics",
    url = "https://aclanthology.org/2024.findings-emnlp.108/",
    doi = "10.18653/v1/2024.findings-emnlp.108",
    pages = "1950--1976"
}

@inproceedings{smart,
    title = "{SMART}: Simulated Students Aligned with Item Response Theory for Question Difficulty Prediction",
    author = "Scarlatos, Alexander  and
      Fernandez, Nigel  and
      Ormerod, Christopher  and
      Lottridge, Susan  and
      Lan, Andrew",
    editor = "Christodoulopoulos, Christos  and
      Chakraborty, Tanmoy  and
      Rose, Carolyn  and
      Peng, Violet",
    booktitle = "Proceedings of the 2025 Conference on Empirical Methods in Natural Language Processing",
    month = nov,
    year = "2025",
    address = "Suzhou, China",
    publisher = "Association for Computational Linguistics",
    url = "https://aclanthology.org/2025.emnlp-main.1274/",
    doi = "10.18653/v1/2025.emnlp-main.1274",
    pages = "25071--25094",
    ISBN = "979-8-89176-332-6"
}

@book{imbens2015causal,
  author    = {Guido W. Imbens and Donald B. Rubin},
  title     = {Causal Inference for Statistics, Social, and Biomedical Sciences: An Introduction},
  publisher = {Cambridge University Press},
  year      = {2015},
  doi       = {10.1017/CBO9781139025751},
}

@inproceedings{gsm-ic,
author = {Shi, Freda and Chen, Xinyun and Misra, Kanishka and Scales, Nathan and Dohan, David and Chi, Ed and Sch{\"a}rli, Nathanael and Zhou, Denny},
title = {Large language models can be easily distracted by irrelevant context},
year = {2023},
publisher = {JMLR.org},
booktitle = {Proceedings of the 40th International Conference on Machine Learning},
articleno = {1291},
numpages = {18},
location = {Honolulu, Hawaii, USA},
series = {ICML'23}
}

@inproceedings{gsm-dc-shi-2025,
    title = "How Is {LLM} Reasoning Distracted by Irrelevant Context? An Analysis Using a Controlled Benchmark",
    author = "Yang, Minglai  and
      Huang, Ethan  and
      Zhang, Liang  and
      Surdeanu, Mihai  and
      Wang, William Yang  and
      Pan, Liangming",
    editor = "Christodoulopoulos, Christos  and
      Chakraborty, Tanmoy  and
      Rose, Carolyn  and
      Peng, Violet",
    booktitle = "Proceedings of the 2025 Conference on Empirical Methods in Natural Language Processing",
    month = nov,
    year = "2025",
    address = "Suzhou, China",
    publisher = "Association for Computational Linguistics",
    url = "https://aclanthology.org/2025.emnlp-main.674/",
    doi = "10.18653/v1/2025.emnlp-main.674",
    pages = "13329--13347",
    ISBN = "979-8-89176-332-6"
}

@article{rubin1974,
author = {Rubin, Donald},
title = {ESTIMATING CAUSAL EFFECTS OF TREATMENTS IN EXPERIMENTAL AND OBSERVATIONAL STUDIES},
journal = {ETS Research Bulletin Series},
volume = {1972},
number = {2},
pages = {i-31},
doi = {https://doi.org/10.1002/j.2333-8504.1972.tb00631.x},
url = {https://onlinelibrary.wiley.com/doi/abs/10.1002/j.2333-8504.1972.tb00631.x},
eprint = {https://onlinelibrary.wiley.com/doi/pdf/10.1002/j.2333-8504.1972.tb00631.x},
year = {1972}
}

@inproceedings{promptbench,
author = {Zhu, Kaijie and Wang, Jindong and Zhou, Jiaheng and Wang, Zichen and Chen, Hao and Wang, Yidong and Yang, Linyi and Ye, Wei and Zhang, Yue and Gong, Neil and Xie, Xing},
title = {PromptRobust: Towards Evaluating the Robustness of Large Language Models on Adversarial Prompts},
year = {2024},
isbn = {9798400712098},
publisher = {Association for Computing Machinery},
address = {New York, NY, USA},
url = {https://doi.org/10.1145/3689217.3690621},
doi = {10.1145/3689217.3690621},
booktitle = {Proceedings of the 1st ACM Workshop on Large AI Systems and Models with Privacy and Safety Analysis},
pages = {57–68},
numpages = {12},
location = {Salt Lake City, UT, USA},
series = {LAMPS '24}
}

@inproceedings{pryzant2021,
    title = "Causal Effects of Linguistic Properties",
    author = "Pryzant, Reid  and
      Card, Dallas  and
      Jurafsky, Dan  and
      Veitch, Victor  and
      Sridhar, Dhanya",
    editor = "Toutanova, Kristina  and
      Rumshisky, Anna  and
      Zettlemoyer, Luke  and
      Hakkani-Tur, Dilek  and
      Beltagy, Iz  and
      Bethard, Steven  and
      Cotterell, Ryan  and
      Chakraborty, Tanmoy  and
      Zhou, Yichao",
    booktitle = "Proceedings of the 2021 Conference of the North American Chapter of the Association for Computational Linguistics: Human Language Technologies",
    month = jun,
    year = "2021",
    address = "Online",
    publisher = "Association for Computational Linguistics",
    url = "https://aclanthology.org/2021.naacl-main.323/",
    doi = "10.18653/v1/2021.naacl-main.323",
    pages = "4095--4109"
}

@inproceedings{veitch2020,
  title={Adapting Text Embeddings for Causal Inference},
  author={Victor Veitch and Dhanya Sridhar and David M. Blei},
  booktitle={Conference on Uncertainty in Artificial Intelligence},
  year={2019},
  url={https://api.semanticscholar.org/CorpusID:220793347}
}

@inproceedings{promptset,
author = {Razavi, Amirhossein and Soltangheis, Mina and Arabzadeh, Negar and Salamat, Sara and Zihayat, Morteza and Bagheri, Ebrahim},
title = {Benchmarking Prompt Sensitivity in Large Language Models},
year = {2025},
isbn = {978-3-031-88713-0},
publisher = {Springer-Verlag},
address = {Berlin, Heidelberg},
url = {https://doi.org/10.1007/978-3-031-88714-7_29},
doi = {10.1007/978-3-031-88714-7_29},
booktitle = {Advances in Information Retrieval: 47th European Conference on Information Retrieval, ECIR 2025, Lucca, Italy, April 6–10, 2025, Proceedings, Part III},
pages = {303–313},
numpages = {11},
location = {Lucca, Italy}
}

@inproceedings{qpp-specificity,
author = {Arabzadeh, Negar and Zarrinkalam, Fattane and Jovanovic, Jelena and Bagheri, Ebrahim},
title = {Neural Embedding-Based Metrics for Pre-retrieval Query Performance Prediction},
year = {2020},
isbn = {978-3-030-45441-8},
publisher = {Springer-Verlag},
address = {Berlin, Heidelberg},
url = {https://doi.org/10.1007/978-3-030-45442-5_10},
doi = {10.1007/978-3-030-45442-5_10},
booktitle = {Advances in Information Retrieval: 42nd European Conference on IR Research, ECIR 2020, Lisbon, Portugal, April 14–17, 2020, Proceedings, Part II},
pages = {78–85},
numpages = {8},
location = {Lisbon, Portugal}
}

@inproceedings{vap3,
author = {Arabzadeh, Negar and Bagheri, Ebrahim},
title = {VAP3: Variation-Aware Prompt Performance Prediction},
year = {2025},
isbn = {9798400715921},
publisher = {Association for Computing Machinery},
address = {New York, NY, USA},
url = {https://doi.org/10.1145/3726302.3730264},
doi = {10.1145/3726302.3730264},
booktitle = {Proceedings of the 48th International ACM SIGIR Conference on Research and Development in Information Retrieval},
pages = {2794–2799},
numpages = {6},
location = {Padua, Italy},
series = {SIGIR '25}
}

@inproceedings{bertpe,
author = {Khodabakhsh, Maryam and Zarrinkalam, Fattane and Arabzadeh, Negar},
title = {BertPE: A BERT-Based Pre-retrieval Estimator for Query Performance Prediction},
year = {2024},
isbn = {978-3-031-56062-0},
publisher = {Springer-Verlag},
address = {Berlin, Heidelberg},
url = {https://doi.org/10.1007/978-3-031-56063-7_27},
doi = {10.1007/978-3-031-56063-7_27},
booktitle = {Advances in Information Retrieval: 46th European Conference on Information Retrieval, ECIR 2024, Glasgow, UK, March 24–28, 2024, Proceedings, Part III},
pages = {354–363},
numpages = {10},
location = {Glasgow, United Kingdom}
}

@inproceedings{gsm-plus,
    title = "{GSM}-Plus: A Comprehensive Benchmark for Evaluating the Robustness of {LLM}s as Mathematical Problem Solvers",
    author = "Li, Qintong  and
      Cui, Leyang  and
      Zhao, Xueliang  and
      Kong, Lingpeng  and
      Bi, Wei",
    editor = "Ku, Lun-Wei  and
      Martins, Andre  and
      Srikumar, Vivek",
    booktitle = "Proceedings of the 62nd Annual Meeting of the Association for Computational Linguistics (Volume 1: Long Papers)",
    month = aug,
    year = "2024",
    address = "Bangkok, Thailand",
    publisher = "Association for Computational Linguistics",
    url = "https://aclanthology.org/2024.acl-long.163/",
    doi = "10.18653/v1/2024.acl-long.163",
    pages = "2961--2984"
}

@article{hendrycks-math,
  author       = {Dan Hendrycks and
                  Collin Burns and
                  Saurav Kadavath and
                  Akul Arora and
                  Steven Basart and
                  Eric Tang and
                  Dawn Song and
                  Jacob Steinhardt},
  title        = {Measuring Mathematical Problem Solving With the {MATH} Dataset},
  journal      = {CoRR},
  volume       = {abs/2103.03874},
  year         = {2021},
  url          = {https://arxiv.org/abs/2103.03874},
  eprinttype   = {arXiv},
  eprint       = {2103.03874},
  bibsource    = {dblp computer science bibliography, https://dblp.org}
}

@article{frangakis2002,
author = {Frangakis, Constantine E. and Rubin, Donald B.},
title = {Principal Stratification in Causal Inference},
journal = {Biometrics},
volume = {58},
number = {1},
pages = {21-29},
doi = {https://doi.org/10.1111/j.0006-341X.2002.00021.x},
url = {https://onlinelibrary.wiley.com/doi/abs/10.1111/j.0006-341X.2002.00021.x},
eprint = {https://onlinelibrary.wiley.com/doi/pdf/10.1111/j.0006-341X.2002.00021.x},
year = {2002}
}

@article{zhang2003,
 ISSN = {10769986, 19351054},
 URL = {http://www.jstor.org/stable/3701340},
 author = {Junni L. Zhang and Donald B. Rubin},
 journal = {Journal of Educational and Behavioral Statistics},
 number = {4},
 pages = {353--368},
 publisher = {[American Educational Research Association, Sage Publications, Inc., American Statistical Association]},
 title = {Estimation of Causal Effects via Principal Stratification When Some Outcomes Are Truncated by "Death"},
 urldate = {2026-09-11},
 volume = {28},
 year = {2003}
}

@inproceedings{amg-rag,
    title = "Agentic Medical Knowledge Graphs Enhance Medical Question Answering: Bridging the Gap Between {LLM}s and Evolving Medical Knowledge",
    author = "Rezaei, Mohammad R.  and
      Fard, Reza Saadati  and
      Parker, Jayson L.  and
      Krishnan, Rahul G.  and
      Lankarany, Milad",
    editor = "Christodoulopoulos, Christos  and
      Chakraborty, Tanmoy  and
      Rose, Carolyn  and
      Peng, Violet",
    booktitle = "Findings of the {A}ssociation for {C}omputational {L}inguistics: {EMNLP} 2025",
    month = nov,
    year = "2025",
    address = "Suzhou, China",
    publisher = "Association for Computational Linguistics",
    url = "https://aclanthology.org/2025.findings-emnlp.679/",
    doi = "10.18653/v1/2025.findings-emnlp.679",
    pages = "12682--12701",
    ISBN = "979-8-89176-335-7"
}

@misc{maricl,
      title={From Residuals to Reasons: LLM-Guided Mechanism Inference from Tabular Data}, 
      author={Mohammad R. Rezaei and Rahul G. Krishnan},
      year={2026},
      eprint={2605.22897},
      archivePrefix={arXiv},
      primaryClass={cs.LG},
      url={https://arxiv.org/abs/2605.22897}, 
}

@inproceedings{vendi-rag,
    title = "Vendi-{RAG}: Adaptively Trading-Off Diversity And Quality Significantly Improves Retrieval Augmented Generation With {LLM}s",
    author = "Rezaei, Mohammad Reza  and
      Dieng, Adji Bousso",
    editor = "Lee, Hung-Yi  and
      Dev, Sunipa  and
      Li, Jessy  and
      Elliott, Desmond",
    booktitle = "Findings of the {A}ssociation for {C}omputational {L}inguistics: {EMNLP} 2026",
    month = oct,
    year = "2026",
    address = "Budapest, Hungary",
    publisher = "Association for Computational Linguistics",
    url = "https://aclanthology.org/2026.findings-emnlp.1499/",
    pages = "31909--31932",
    ISBN = "979-8-89176-462-0"
}
\bibliographystyle{iclr2027_conference}

\appendix

\section{Proof of Proposition~\ref{prop:orthogonality}}
\label{app:proof}

Holds for either $Q\in\{\tau,\pi\}$; write $Q$ throughout. (i) $\gamma(C)=\E[Q\mid C]$ is the $L_2$ projection of $Q$ onto $L_2(\sigma(C))$, so the residual $D^{\perp}=Q-\gamma(C)$ is orthogonal to that subspace: taking $\eta\equiv1$ gives $\E[D^{\perp}]=0$, and for general $\eta\in L_2(\sigma(C))$, $\mathrm{Cov}(D^{\perp},\eta(C))=\E[D^{\perp}\eta(C)]=0$. Note that the projection is onto functions of the \emph{declared} $C$ (residuals of a declared regression are likewise analyzed in \citealt{maricl}). (ii) For the plug-in estimator, $\hat D^{\perp}-D^{\perp}=(\hat Q-Q)-(\hat\gamma-\gamma)$; both terms vanish in $L_2$ as $\hat Q\to Q$ ($\hat\pi\to\pi$ by consistency of the binomial rate as the number of observations grows; $\hat\tau\to\tau$ by per-job concentration as $G\to\infty$) and the cross-fitted complexity regression is consistent, removing own-observation leakage.

\section{Synthetic data (Synth-CD)}
\label{app:synthetic}

A synthetic suite with \emph{two} independent
ground-truth factors: structural complexity $C\sim\mathrm{Beta}(2,2)$ and domain expertise $E\sim\mathrm{Beta}(2,2)$ ($C\perp E$ by construction). Eight
domains (web development, data engineering, ML operations, scientific writing, legal review, medical coding, financial modeling, creative content), each with 12 operator types spanning four semantic tiers from cosmetic-only through dual-sensitive. Text length is controlled to be independent of $C$
and $E$, preventing the surface-length heuristic from proxying either factor. 1{,}000 jobs, $G{=}12$ twins. Scores here are continuous and per-twin effects are non-negative by construction, so the UCD contains no repairs and measures degradation directly on this suite (on the real suites it also nets in repairs, \S\ref{sec:new-estimand}); we use it accordingly as a mechanism check rather than as a headline setting.

\subsection{Synth-CD: dual-factor robustness check}
\label{app:synth2}

\textbf{Generating process.} The per-twin effect is
$\delta(J,\mathrm{op})=\max\!\bigl(0,\ \mu_{\mathrm{op}}(1{+}C(J))
+\zeta_{\mathrm{op}}\,E(J)+\varepsilon\bigr)$, calibrated so that
$\mathrm{Spearman}(D_{\mathrm{avg}},C)\approx\mathrm{Spearman}(D_{\mathrm{avg}},E)\approx0.60$
and $\mathrm{Spearman}(C,E)\approx0.03$ (near-zero) in the holdout; the oracle $D^{\perp}(J)=D_{\mathrm{avg}}(J)-\gamma(C(J))$ concentrates on the expertise signal ($\mathrm{Spearman}(\text{oracle }D^{\perp},E)\approx0.75$).

\begin{table}[ht]
\centering\small
\caption{Synth-CD: $D_{\mathrm{avg}}$ prediction (1{,}000 jobs, $G{=}12$ twins, 300 held-out, MiniLM backbone). Text length is controlled to be independent of $C$ and $E$. Score-predictor: predicts the base score $Y(J)$ from text and uses it as a difficulty score. $\rho_S$, NDCG@10, AUROC~$\uparrow$; MAE, $|\rho_C|$ shown for context.}
\label{tab:synth2}
\begin{tabular}{@{}lccccc@{}}
\toprule
Method & $\rho_S\uparrow$ & MAE$\downarrow$ & NDCG@10$\uparrow$ & AUROC$\uparrow$ & $|\rho_C|$ \\
\midrule
Surface length          & 0.100 & 0.021 & 0.804 & 0.594 & 0.052 \\
Score-predictor         & 0.496 & 0.014 & 0.829 & 0.727 & 0.052 \\
Frozen twin-distance & $-$0.112 & 0.023 & 0.686 & 0.446 & 0.182 \\
\FrailTau\ ex-ante (ours)      & 0.420 & 0.015 & 0.799 & 0.712 & 0.067 \\
\textbf{\FrailTau\ contrast (ours)} & \textbf{0.617} & \textbf{0.012} & \textbf{0.863} & \textbf{0.828} & \textbf{0.028} \\
\bottomrule
\end{tabular}
\end{table}

Table~\ref{tab:synth2} shows the \FrailTau\ contrast head reaching $\rho_S{=}0.617$ and AUROC~$0.828$, beating the
score-predictor by $0.12$ Spearman points and the ex-ante head by $0.20$; surface length scores near-random ($\rho_S{=}0.100$), confirming the
advantage is semantic rather than a token-count artifact. The contrast head also reaches the lowest $|\rho_C|$ of any predictor ($0.028$) \emph{without}
residualization. The deployable ex-ante head, which sees no twin, trails the score-predictor here ($0.420$ vs.\ $0.496$), so on Synth-CD the advantage belongs to the contrast head.

\section{Implementation details}
\label{app:impl}

\textbf{Two-stage curriculum.} Every \FrailPi/\FrailTau\ model, frozen or fine-tuned, trains in two stages: a label-free representation stage (triplet loss pulling twins
together, pushing requirement-changing edits apart) and a score-labeled calibration stage via \eqref{eq:loss}. For the frozen-backbone experiments, we employ an off-the-shelf small foundation model as the encoder, freeze the backbone, and learn only the heads.

\textbf{Frozen-backbone runs (Synth-CD and the frozen ablations).} Synth-CD predictor heads: hidden dim 256, complexity dim 64, dropout 0.15; contrast MLP 3 layers, ex-ante MLP 4 layers. Stage~1 (representation): 150 epochs, triplet margin 0.5, lr~$10^{-3}$, batch 128. Stage~2 (calibration): 400 epochs, lr~$3{\times}10^{-4}$, AdamW, weight decay~$10^{-4}$, $\lambda_{\mathrm{inv}}{=}1.5$, $\delta_{\mathrm{Huber}}{=}0.05$ (Synth-CD; the real-suite frozen ablations use cross-entropy, as the main model does), cosine LR decay, batch 128. Synth-CD uses a
frozen \texttt{all-MiniLM-L6-v2} (384-dim) with pre-computed, cached embeddings (${\approx}2$ min on an M-series CPU). The GlassGSM/GlassMATH frozen ablations of Table~\ref{tab:c1-repr} use a frozen \texttt{text-embedding-3-small} (the ``Frozen OpenAI embedding'' row, 1536-dim) and a frozen \texttt{BAAI/bge-small-en-v1.5} (the ``Frozen sentence encoder'' row, 384-dim); their heads instead use hidden dim 64, complexity dim 32/64, dropout 0.25, a 2-layer ex-ante head (avoiding overfitting the modest per-suite job
count).

\textbf{Main model (\FrailPi, fine-tuned encoder).} The headline real-suite model of \S\ref{sec:c1} instead fine-tunes BERT-base (768-dim matching BERTPE's own default so the comparison is
like-for-like) end to end. Heads: hidden dim 256, complexity dim 32, dropout 0.25, 2-layer ex-ante
head. Stage~1 (representation): 10 epochs, triplet margin 0.5, head
lr~$10^{-3}$, encoder lr~$2{\times}10^{-5}$, batch 128. Stage~2 (calibration):
20 epochs, head lr~$5{\times}10^{-4}$, encoder lr~$2{\times}10^{-5}$, AdamW,
weight decay~$10^{-2}$, binary cross-entropy loss, batch 128. The much
lower encoder lr keeps full fine-tuning from washing out the pretrained
representation in the few epochs the labelled pair count supports.

\subsection{Extended baselines}
\label{app:extended-baselines}

Here we detail how five published sensitivity predictors, plus our own frozen-geometry baseline, are adapted to this task: BERTPE~\citep{bertpe}
and VAP3~\citep{vap3} require retraining (VAP3 also reads a twin at evaluation time, through its pair encoder); Frozen twin-distance, Prompt Clarity, and Embedding Entropy~\citep{promptset} need a job's twins at evaluation time; Embedding Specificity-IEF~\citep{promptset,qpp-specificity} is ex-ante (base text only, scored against the training pool). Unlike BERTPE/VAP3,
none of the last four fine-tunes a representation: each pairs a frozen TF-IDF+SVD embedding (the same representation Table~\ref{tab:c1-repr} ablates for \FrailPi) with the method's own unsupervised signal (twin-embedding
distance, twin cosine similarity, pairwise-similarity entropy, or corpus specificity), and only a final linear calibration is fit to $\pi$ --- retargeting
changes what the calibration maps to, not what the signal computes. All four score near or below zero on both suites (Table~\ref{tab:c1}: $\le0.060$ of ceiling in absolute value on GlassGSM, $\le0.160$ on GlassMATH): their unsupervised signals carry little to no rank information about which solvable jobs a model's twins will
break.

\textbf{Adaptation of supervised baselines.} BERTPE and VAP3 were originally
trained for binary per-prompt correctness classification via a binary cross-entropy (BCE) loss. We preserve their published architectures and hyperparameters verbatim --- BERTPE's single-tower BERT$\to$CLS$\to$MLP head and VAP3's concatenated pair-encoder (base$\oplus$[SEP]$\oplus$twin) with FGSM adversarial perturbations on token embeddings (equations 5--8 of \citealt{vap3}, $\varepsilon{=}0.01$, $\lambda{=}0.5$) --- and replace only the loss: BERTPE regresses $\pi$ directly (MSE, one sample per base job, prediction is the calibrated MLP output); VAP3 regresses the per-observation break event $B(J,T)=\mathbf{1}\{Y(J){=}1,\,Y(T(J)){=}0\}$ (MSE, one sample per pair), with job-level difficulty the mean-over-twins aggregation of calibrated pair predictions, since $\pi$ is a rate, not a worst case. Both models are fine-tuned from BERT-base with AdamW.

\subsection{C1 ablations: representation and training recipe}
\label{app:c1-ablations}

Other variants of \FrailPi: the text-representation comparison and five further training-recipe levers (the frozen-encoder variant of the cross-generator check is shown in Figure~\ref{fig:cross-model}). Table~\ref{tab:c1-repr} compares the main model's fine-tuned BERT-base encoder (to isolate the representation) against three frozen/non-adaptive representations. Neither frozen encoder beats plain TF-IDF, so text representation alone is not the bottleneck; letting the encoder \emph{adapt} to the break-probability objective is what yields a real gain over every frozen representation, on GlassGSM ($0.613$ vs.\ a best frozen $0.398$) and on GlassMATH ($0.557$ vs.\ $0.435$) alike.

\begin{table}[ht]
\centering\small
\caption{Representation ablation for \FrailPi: test-fold Spearman
\emph{normalized by the suite reliability ceiling} ($0.969$ for GlassGSM,
$0.972$ for GlassMATH; Appendix~\ref{app:audit}), mean${\pm}$sd over five seeds;
$1.0$ is the attainable maximum. Same protocol and normalization as
Table~\ref{tab:c1}, so these rows are directly comparable to the main model's
$0.613$/$0.557$.}
\label{tab:c1-repr}
\begin{tabular}{@{}llccc@{}}
\toprule
Suite & Representation & $\pi$ $\uparrow$ & $a$ $\uparrow$ & $b$ $\uparrow$ \\
\midrule
\multirow{3}{*}{GlassGSM}
 & TF-IDF+SVD                     & 0.398${\pm}$0.042 & 0.395${\pm}$0.037 & 0.039${\pm}$0.070 \\
 & Frozen sentence encoder        & 0.381${\pm}$0.026 & 0.381${\pm}$0.032 & $-$0.126${\pm}$0.112 \\
 & Frozen OpenAI embedding        & 0.326${\pm}$0.040 & 0.324${\pm}$0.039 & $-$0.158${\pm}$0.046 \\
\midrule
\multirow{3}{*}{GlassMATH}
 & TF-IDF+SVD                     & 0.435${\pm}$0.021 & 0.425${\pm}$0.021 & 0.334${\pm}$0.095 \\
 & Frozen sentence encoder        & 0.386${\pm}$0.013 & 0.394${\pm}$0.014 & 0.314${\pm}$0.037 \\
 & Frozen OpenAI embedding        & 0.431${\pm}$0.016 & 0.417${\pm}$0.015 & 0.369${\pm}$0.022 \\
\bottomrule
\end{tabular}
\end{table}

Table~\ref{tab:c1-ablations} tries five further levers against the main model; none beats it meaningfully. A stronger backbone (RoBERTa-large in place of \texttt{Bert-base}) trails on GlassGSM and is within seed noise on GlassMATH. Lower LR / higher patience is worse on GlassGSM and flat on GlassMATH; a larger job batch is within seed noise on GlassGSM (not run on GlassMATH). Cross-representation ensembling and dataset pooling are worse on GlassGSM and marginally (${\le}0.009$) below the main model on GlassMATH, within seed noise.

\begin{table}[ht]
\centering\small
\caption{Further training-recipe levers vs.\ the $0.613$/$0.557$ main model
(break rate $\rho_S$ \emph{normalized by the suite reliability ceiling}, $0.969$/$0.972$; Appendix~\ref{app:audit};
same normalization as Table~\ref{tab:c1}, $1.0$ is the attainable maximum; 5 seeds unless noted). None beats
the main model meaningfully.}
\label{tab:c1-ablations}
\begin{tabular}{@{}lcc@{}}
\toprule
Lever & GlassGSM & GlassMATH \\
\midrule
Main model (fine-tuned encoder) & 0.613 & 0.557 \\
Stronger backbone (RoBERTa-large) & 0.543 & 0.548 \\
Lower LR / higher patience & 0.567 & 0.552 \\
Larger job batch & 0.606 & --- \\
Ensemble (mean-rank / mean-z) & 0.559 / 0.537 & 0.549 / 0.551 \\
Pooled training data (per-origin)$^\S$ & 0.536 & 0.548 \\
\bottomrule
\end{tabular}\\[4pt]
{\footnotesize $^\S$Trained on GlassGSM and GlassMATH pooled, then scored
separately on each suite's own holdout. ``---'': lever not run on that suite.}
\end{table}

\subsection{Real-suite UCD baselines: why the unconditional target fails}
\label{app:c1-davg-real}

This is the empirical basis for the estimand choice of \S\ref{sec:new-estimand}. Table~\ref{tab:c1-davg-real} complements the Synth-CD $D_{\mathrm{avg}}$ comparison (Table~\ref{tab:synth2}) with a broader baseline sweep on a real suite, 5 seeds, scored directly against the UCD rather than the CCD. Raw $D_{\mathrm{avg}}$ is itself correlated with the declared complexity proxy $C$ (Spearman$(D_{\mathrm{avg}},C)=0.278{\pm}0.040$), so, unlike the CRD results in Table~\ref{tab:c2}, no method here should be read as complexity-orthogonal. Every predictor is weak: the best $\rho_S$ is surface length's $0.162{\pm}0.030$, and \FrailTau\ reaches the lowest MAE ($0.108{\pm}0.016$) but only $\rho_S{=}0.092$. Against \FrailPi's $0.594$ on the solvable stratum of the same corpus, this is the difference the conditioning makes: on real, binary-scored suites the unconditional average nets breaks on solved tasks against repairs on unsolved ones and is a noisy target for \emph{every} method, which is why the paper's claims are stated for the per-observation CCD/CRD estimands.

\begin{table}[ht]
\centering
\caption{GlassGSM (real suite): UCD prediction on the audited, multi-operator-mix, pooled holdout, mean${\pm}$sd over 5 seeds. Correlational score-predictor: predicts the base score $Y(J)$ from text and uses it as a difficulty score; Embedding linear: RidgeCV on frozen embeddings regressed on $D_{\mathrm{avg}}$. $\rho_S$, NDCG@10, AUROC~$\uparrow$; MAE, $|\rho_C|$ shown for context. }
\label{tab:c1-davg-real}
\resizebox{\textwidth}{!}{
\begin{tabular}{@{}lccccc@{}}
\toprule
Method & $\rho_S\uparrow$ & MAE$\downarrow$ & NDCG@10$\uparrow$ & AUROC$\uparrow$ & $|\rho_C|$ \\
\midrule
Surface length                   & \textbf{0.162${\pm}$0.030} & 0.245${\pm}$0.009 & \textbf{0.453${\pm}$0.065} & \textbf{0.727${\pm}$0.041} & 0.086${\pm}$0.040 \\
Correlational score-predictor    & 0.028${\pm}$0.036 & 0.254${\pm}$0.010 & 0.030${\pm}$0.036 & 0.544${\pm}$0.045 & 0.037${\pm}$0.019 \\
Frozen twin-distance & $-$0.101${\pm}$0.054 & 0.238${\pm}$0.012 & 0.107${\pm}$0.075 & 0.366${\pm}$0.044 & 0.065${\pm}$0.051 \\
BERTPE                            & 0.073${\pm}$0.043 & 0.192${\pm}$0.010 & 0.175${\pm}$0.140 & 0.611${\pm}$0.052 & 0.044${\pm}$0.029 \\
VAP3                              & 0.092${\pm}$0.089 & 0.189${\pm}$0.034 & 0.186${\pm}$0.168 & 0.621${\pm}$0.095 & 0.059${\pm}$0.042 \\
Embedding Specificity-IEF (PromptSET) & 0.022${\pm}$0.005 & 6.102${\pm}$3.501 & $-$0.011${\pm}$0.134 & 0.508${\pm}$0.002 & 0.030${\pm}$0.005 \\
Embedding Entropy (PromptSET)    & 0.011${\pm}$0.026 & 0.214${\pm}$0.011 & $-$0.013${\pm}$0.027 & 0.529${\pm}$0.036 & 0.091${\pm}$0.039 \\
Embedding linear (RidgeCV) & 0.053${\pm}$0.064 & 0.170${\pm}$0.006 & 0.079${\pm}$0.104 & 0.580${\pm}$0.061 & \textbf{0.024${\pm}$0.011} \\
Prompt Clarity (PromptSET)       & $-$0.106${\pm}$0.051 & 0.188${\pm}$0.010 & 0.118${\pm}$0.088 & 0.355${\pm}$0.045 & 0.066${\pm}$0.042 \\
\textbf{\FrailTau\ (ours)}      & 0.092${\pm}$0.059 & \textbf{0.108${\pm}$0.016} & 0.197${\pm}$0.122 & 0.614${\pm}$0.051 & 0.040${\pm}$0.021 \\
\bottomrule
\end{tabular}
}
\end{table}

\section{Benchmarks}
\label{app:benchmarks}

Construction, statistics, and audits behind every
real-data benchmark used in the paper: the datasets themselves, why the public alternatives fall short of R1--R3 (\S\ref{sec:benchmarks}), the measurement audit (reliability ceiling and identity-twin floor),
the empirical No-Op audit, and the downstream-utility ranking check.

\subsection{Datasets}
\label{app:datasets}

\textbf{GSM-Symbolic (not used in main experiments).} The GSM-Symbolic p1 set~\citep{gsm-symbolic} changes numbers and names, which changes the correct answer and violates the No-Op requirement (Assumption~\ref{ass:invariance}); we exclude it (see \S\ref{sec:benchmarks} and Table~\ref{tab:fragility-comparison}).

\textbf{GSM-IC (irrelevant context).} GSM8K problems with a single injected irrelevant clause. The released GSM-IC~\citep{gsm-ic} has 100 base problems; we apply its construction to all 1{,}319 GSM8K test problems (20 twins each), the base set GlassGSM also uses; trivially-ignorable distractors produce near-zero fragility on current LLMs (the composition of supplied context matters more in knowledge-grounded and retrieval-augmented QA; \citealt{amg-rag,vendi-rag}).

\textbf{MATH (original).} The Hendrycks MATH benchmark~\citep{hendrycks-math} (\texttt{EleutherAI/hendrycks\_math}, test split): $5{,}000$ competition-level problems across seven subjects, each with a difficulty level (1--5) mapped to \texttt{complexity\_score}. No perturbation suite is provided (no twins). Its level supplies a non-circular $C$ proxy, but without twins neither fragility (\textbf{R2}) nor its relation to complexity (\textbf{R3}) can be measured.

\textbf{GlassGSM (ours).} An upgraded GSM-IC whose twins are built from three declared MPPO families (entity renaming, sentence reordering, and injecting a math-domain setup sentence from a \emph{different} GSM8K problem) raising fragility from $3.7\%$ to $44.9\%$ over the pooled doses while preserving the answer key. Twins are generated across all three families at each dose $m\in\{1,3,5,7,9\}$; higher doses inject more distractor sentences. We read $m{=}1$ as the conceptually clean case (Figure~\ref{fig:overview}(a) shows a single reordering). The released corpus covers all 1{,}319 problems; its five dose files, resampled from the audited pools, hold $32{,}858$ pairs and $18{,}309$ unique twins ($13.9$ per problem, deduplicated on twin text).

\textbf{GlassMATH (ours).} The GlassGSM protocol applied to Hendrycks MATH: base~$=$~original problem, twin~$=$~the same problem perturbed with the same three families and doses while preserving the gold answer. Each problem carries its MATH difficulty level and subject tag; level maps to an external \texttt{complexity\_score} in $[0.05,0.95]$, a non-circular complexity proxy. GlassMATH carries 7 MATH subject categories (Algebra, Prealgebra, Number Theory, Intermediate Algebra, Geometry, Counting \& Probability, Precalculus). Like GlassGSM, it has five dose files ($m\in\{1,3,5,7,9\}$), each resampled from a pool that passed the same audit; together they hold $122{,}847$ pairs and $115{,}364$ unique twins over $4{,}999$ problems ($23.1$ per problem; Table~\ref{tab:fragility-comparison}), and the audited $m{=}9$ pool is the $94{,}750$ pairs of Table~\ref{tab:noop-audit}.

\subsection{Measurement audit: what the targets actually are}
\label{app:audit}
A parametric bootstrap --- with a cross-validated empirical-Bayes prior (fit on held-out job folds, so no job's own rate informs the prior it is shrunk toward) and replicates drawn from each fold's posterior rather than a fixed
shrunk point estimate, to avoid overstating agreement from shared shrinkage --- agrees closely with the classical split-half estimator ($0.939$ vs.\ $0.941$ on GlassGSM, $0.945$ vs.\ $0.951$ on GlassMATH); taking $r$ as the bootstrap estimate gives an implied ceiling on
any predictor's Spearman of $\sqrt{r}{=}0.969$ and $0.972$ respectively.

\paragraph{Identity-twin floor.}
\label{app:placebo}
Because solver outputs are sampled, $B(J,T)$ can fire even when $T$ has no effect. We measure this floor with identity twins, $T(J)=J$: the unperturbed base prompts of solvable-stratum items that the reference ($m{=}9$) run also solves were re-solved in independent runs of the same pipeline, solver, and decoding (temperature $0.2$, $5$ samples, modal answer; four further runs per item on GlassGSM, three on GlassMATH), and a re-solve counts as a break when its extracted answer differs from the reference run's correct answer. Selecting on one run and scoring the others mirrors how $B$ is computed, so the floor also absorbs any regression to the mean from conditioning on a measured $Y(J)$; it is the average CCD that a perturbation family with no effect would receive on the same stratum. Table~\ref{tab:placebo} reports it. The floor is small on GlassGSM and larger on GlassMATH; on both suites the modal base answer almost never moves, so an item's stratum membership is reproducible rather than the outcome of one lucky draw.

\begin{table}[ht]
\centering\small
\caption{Identity-twin floor: break rate of re-solved, unperturbed base prompts on the solvable stratum (the average CCD of a no-effect perturbation).}
\label{tab:placebo}
\begin{tabular}{@{}lccc@{}}
\toprule
Suite & Re-runs/item & Floor & Modal answer $\neq$ ref. \\
\midrule
GlassGSM &  4 & $1.1\%$ & $4$ ($0.3\%$) \\
GlassMATH   &  3 & $2.5\%$ & $0$ ($0.0\%$) \\
\bottomrule
\end{tabular}
\end{table}

\subsection{Downstream utility: top-k curation}
\label{app:utility}

C1--C2 are measured against the estimand; a practitioner cares whether the ranking is actionable. We run the curation experiment the method is motivated by: on the audited GlassGSM holdout, select the top-$k\%$ of jobs by each predictor ($k\in\{5,10,20,50\}$) and report recall of truly
fragile jobs ($\pi>0.05$) against Surface features (GBT) and Declared complexity --- the two simple baselines of Table~\ref{tab:c1} --- and an oracle ranking by the measured $\pi$ itself. \FrailPi\ here is the real-suite headline predictor (\S\ref{sec:c1}), ranked by its actual held-out predictions from the audited, pooled 5-dose run behind Table~\ref{tab:c1} (same 5 seeds and splits, no retraining).

\begin{table}[ht]
\centering
\caption{Recall of truly fragile jobs ($\pi>0.05$) in the top-$k\%$ of the GlassGSM holdout ranked by each predictor, mean${\pm}$sd over the same 5 seeds as Table~\ref{tab:c1} (373 held-out jobs/seed, $50.8\%$ fragile on average). Random is the expected recall $k$; oracle ranks by the measured $\pi$ itself. The practically meaningful quantity is the lift over random at small $k$.}
\label{tab:utility}
\resizebox{\textwidth}{!}{
\begin{tabular}{@{}lcccc@{}}
\toprule
Ranking & Recall@5\% & Recall@10\% & Recall@20\% & Recall@50\% \\
\midrule
Random (expected)          & 0.05 & 0.10 & 0.20 & 0.50 \\
Surface features (GBT)     & 0.066${\pm}$0.005 & 0.135${\pm}$0.012 & 0.251${\pm}$0.026 & 0.555${\pm}$0.044 \\
Declared complexity        & 0.064${\pm}$0.007 & 0.122${\pm}$0.006 & 0.244${\pm}$0.007 & 0.567${\pm}$0.018 \\
\textbf{\FrailPi\ (ours)}           & \textbf{0.099${\pm}$0.002} & \textbf{0.189${\pm}$0.005} & \textbf{0.359${\pm}$0.015} & \textbf{0.725${\pm}$0.020} \\
$\widehat D^{\perp}$ (ours) & 0.099${\pm}$0.002 & 0.186${\pm}$0.006 & 0.352${\pm}$0.021 & 0.676${\pm}$0.020 \\
Oracle ($\pi$)              & 0.100${\pm}$0.001 & 0.195${\pm}$0.002 & 0.396${\pm}$0.004 & 0.981${\pm}$0.011 \\
\bottomrule
\end{tabular}
}
\end{table}
On the audited GlassGSM holdout (Table~\ref{tab:utility}), \FrailPi\ roughly doubles random recall at small $k$ (Recall@5\%${=}0.099$ vs.\ $0.05$, matching the oracle's $0.100$ almost exactly; with $50.8\%$ of jobs fragile, ${\approx}0.10$ is the most any ranking can reach at $k{=}5\%$) and stays ahead of both simple baselines at every budget; Surface features and Declared complexity each beat random by a modest, similar margin, well below \FrailPi. The CRD ranking ties raw $\pi$ at Recall@5\% and trails it slightly at larger budgets (e.g.\ $0.676$ vs.\ $0.725$ at Recall@50\%) --- the expected cost of trading
some raw ranking power for orthogonality to the declared complexity proxy.
Evaluating the top $20\%$ of the corpus surfaces ${\approx}36\%$ of fragile items, against $20\%$ for random and $39.6\%$ for the oracle.

\subsection{Empirical No-Op and collision audit of the generated twins}
\label{app:noop-audit}

Here we check whether a twin's perturbations preserve the gold answer (the No-Op requirement, R1) rather than mislabeling a scoring artifact as causal difficulty. Twins are LLM-authored, applying operators from three families (entity renaming, sentence reordering, injection); since an LLM writes every twin, any family, not only injection, can alter answer-relevant text, so the audit covers all three families. Both real-math
suites (\texttt{GlassGSM}, \texttt{GlassMATH}) were passed through a full-corpus audit: every (base, twin)
pair is adjudicated, and a pair is dropped before it enters any benchmark if \textbf{any} of the following holds.

\textbf{LLM No-Op adjudication.} An LLM adjudicator receives the base text, the twin text, and the gold answer, and judges whether the twin's perturbations change or invalidate the correct answer. A ``contradiction'' verdict, or an unparsable response after retries, drops the pair.

\textbf{Renaming and reordering checks.} The two non-injection families are checked directly: a renamed entity must keep its gender and pronoun references (he/she/it), so renaming cannot introduce a referential error that changes the answer, and a reordered twin must preserve the base problem's gold answer.

\textbf{Structural collision checks.} Deterministic local re-implementations of the pipeline's No-Op structural audit catch contamination modes that an LLM judge alone can miss:
\begin{itemize}
    \item \emph{Entity-name collision}: the injected distractor reuses a proper noun from the target problem, so referential ambiguity is introduced (calendar words such as weekday/month names are excluded).
    \item \emph{Gold-answer-number collision}: the injected text contains the numeric value of the correct answer, so a model quoting the distractor number is scored correct for the wrong reason.
    \item \emph{Multi-question collision}: an injected distractor containing its own ``?'' competes for attention, violating single-question
    No-Op.
    \item \emph{Unit-compatible quantity collision}: a shared numeric quantity with a recognized unit. This heuristic flag on GlassMATH alone would flag ${\approx}10$K of the ${\approx}100$K pairs, yet re-adjudicating \emph{all} ${\approx}10$K such flagged-but-kept pairs with an independent LLM adjudicator (Claude, same prompt as the contradiction check above) returned no actual contradiction --- the only borderline flags traced to a pre-existing data gap (the stored problem text omitted its own multiple-choice answer options, making the lettered gold answer unverifiable independent of the injected text), not a semantic effect of the distractor. So it is recorded but, conservatively, not used to drop pairs.
\end{itemize}

Any entity-, gold-answer-, or multi-question collision, and any LLM contradiction verdict, removes the pair; the surviving pairs constitute the benchmarks used throughout the paper. Table~\ref{tab:noop-audit} reports the resulting counts for both suites; the renaming/reordering and multi-question checks removed no pairs, so it has no row for them.

\textbf{Human-in-the-loop audit.} To complement the automated filter, we audited a $1{,}000$-pair sample, drawn from both suites, spanning every audit outcome above (LLM-flagged contradictions, unparsed adjudicator responses, unit-collision pairs flagged but kept, and unflagged pairs) with two independent LLM adjudicators --- \texttt{gpt-5}, the pipeline's own adjudicator, and Claude, queried separately --- followed by human review of the full sample against the base and twin text, the gold answer, and both models' rationales. The audit confirms the filter errs conservatively rather than laxly: the sampled \texttt{llm\_contradiction} drops trace to the adjudicator re-deriving the \emph{original} problem's answer and making its own arithmetic slip, not to any semantic effect of the perturbation, so borderline pairs are excluded rather than risking contamination reaching the benchmark. The unit-collision-flagged-but-kept pairs were independently confirmed clean by the second adjudicator over the full ${\approx}10$K flagged pool, and by both LLM judges and the human reviewer on the sampled subset, with the human and LLM verdicts agreeing on every audited kept pair. The LLM cross-check is therefore full-corpus over the flagged pool; the human pass remains a $1{,}000$-pair spot-check rather than a full-corpus human audit, but at that scale it agrees with both LLM judges on every kept pair.

\begin{table}[ht]
\centering\small
\caption{Contamination audit of the two real-math suites. ``Kept'' is the number of pairs entering
the audited pool from which the benchmarks are drawn (both columns are the $m{=}9$ generation files; the other dose files pass the same audit. The released
corpora resample at most nine of these audited twins per problem at $m{=}9$ and pool the five doses for Table~\ref{tab:fragility-comparison}); ``before'' is the raw un-audited count. The reason rows are not mutually exclusive ---
11 GlassGSM pairs and 62 GlassMATH pairs are dropped for two reasons at once, so each reason row
counts them again and the reasons sum to slightly more than \emph{Dropped}.}
\label{tab:noop-audit}
\begin{tabular}{lcc}
\toprule
 & \texttt{GlassMATH} & \texttt{GlassGSM} \\
\midrule
Pairs before audit & 107{,}573 & 39{,}211 \\
\emph{Kept (audited pool)} & \textbf{94{,}750} & \textbf{34{,}353} \\
Dropped & 12{,}823 & 4{,}858 \\
\hspace{1em}gold-answer collision & 12{,}059 & 4{,}738 \\
\hspace{1em}entity collision & 608 & 35 \\
\hspace{1em}LLM contradiction & 166 & 95 \\
\hspace{1em}unparsed response & 52 & 1 \\
Unique problems & 4{,}992 & 1{,}319 \\
\bottomrule
\end{tabular}
\end{table}

The drop rates are comparable (12.4\% for GlassGSM vs.\ 11.9\% for GlassMATH); GlassMATH shows more collisions in absolute terms because it has ${\approx}2.7\times$ more pairs (107{,}573 vs.\ 39{,}211, from 5{,}000 vs.\ 1{,}319 base
problems), with gold-answer collisions dominating both. Both suites draw distractor sentences from the same GSM8K, non-question setup source (neither contains question-bearing filler), so the heavier absolute footprint on GlassMATH reflects its greater scale and the numeric density of GSM8K filler colliding with the target's own gold value, not a distinct filler type. Both suites use LLM-authored twins built from all three operator families (entity renaming, sentence reordering, and injection), so this audit (which adjudicates every pair across all three families, since an LLM-authored rewrite of any family can alter answer-relevant text) is the sole gate before the targets of \S\ref{sec:new-estimand} are measured.

\end{document}